\documentclass{article}

\PassOptionsToPackage{numbers, compress}{natbib}

\usepackage[final]{neurips_2021}

\makeatletter
\usepackage{microtype}
\usepackage{mdframed}
\usepackage{graphicx}
\usepackage{enumitem}
\usepackage{booktabs} 

\usepackage{color}
\usepackage[usenames,dvipsnames]{xcolor}
\usepackage{xspace}
\usepackage{verbatim}
\usepackage{tikz}
\usepackage{latexsym}
\usepackage{titlesec}
\usepackage{listings}
\usepackage{courier}
\usepackage[T1]{fontenc}

\usepackage{amsmath}
\usepackage{amssymb}

\usepackage{subcaption}
\usepackage{caption}
\usepackage{multirow}
\usepackage{amsfonts, amsmath}
\usepackage{subcaption}
\usepackage{soul}
\usepackage{bm}
\usepackage{wrapfig}
\usepackage{csquotes}

\usepackage{hyperref}       

\newcommand{\whilec}{\textbf{while}}

\newcommand{\ifc}{\textbf{if}}

\newcommand{\thenc}{\textbf{then}}
\newcommand{\elsec}{\textbf{else}}

\newcommand{\id}[1]{\mathsf{#1}}
\newcommand{\Prog}{\mathit{Prog}}

\newcommand{\sY}{\mathsf{Y}}
\newcommand{\sX}{\mathsf{X}}

\newcommand{\sZ}{\mathsf{Z}}

\newcommand{\eat}[1]{} 

\newcommand{\xp}[2]{P \if*#1\else^{#1}\fi \if*#2\else_{\! #2}\fi}

\lstnewenvironment{codesmall2}
    {\lstset{
            aboveskip=5pt,
            belowskip=5pt,
            escapechar=!,
            mathescape=true,
            language=Java,
            basicstyle=\linespread{0.94}\ttfamily\small,
            morekeywords={createSet, sendData, makeLambdaFromMember,
                    APPLY, FILTER, execute, setInput, setOutput, makeLambda, makeLambdaFromMethod,
                    Handle, Vector, Map, pair, makeObjectAllocatorBlock, push_back, Object, makeObject},
            showstringspaces=true}
            \vspace{0pt}%
            \noindent\minipage{0.47\textwidth}}
  {\endminipage\vspace{0pt}}

\lstnewenvironment{codesmall}
    {\lstset{
            aboveskip=5pt,
            belowskip=5pt,
            escapechar=!,
            mathescape=true,
            language=Java,
            basicstyle=\linespread{0.94}\ttfamily\scriptsize,
            morekeywords={createSet, sendData, makeLambdaFromMember,
                    APPLY, FILTER, execute, setInput, setOutput, makeLambda, makeLambdaFromMethod,
                    Handle, Vector, Map, pair, makeObjectAllocatorBlock, push_back, Object, makeObject},
            showstringspaces=true}
            \vspace{0pt}%
            \noindent\minipage{0.47\textwidth}}
  {\endminipage\vspace{0pt}}

\lstnewenvironment{codefootnote}
    {\lstset{
            aboveskip=5pt,
            belowskip=5pt,
            escapechar=!,
            mathescape=true,
            language=Java,
            basicstyle=\linespread{0.94}\ttfamily\scriptsize,
            morekeywords={createSet, sendData, makeLambdaFromMember,
                    APPLY, FILTER, execute, setInput, setOutput, makeLambda, makeLambdaFromMethod,
                    Handle, Vector, Map, pair, makeObjectAllocatorBlock, push_back, Object, makeObject},
            showstringspaces=true}
            \vspace{0pt}%
            \noindent\minipage{0.47\textwidth}}
  {\endminipage\vspace{0pt}}

\newcommand{\nag}{\textsc{Nag}\xspace}
\renewcommand{\nag}{\textsc{Nsg}\xspace}
\newcommand{\codegpt}{\textsc{CodeGPT}\xspace}
\newcommand{\codex}{\textsc{Codex}\xspace}
\newcommand{\gptneo}{\textsc{GPT-Neo}\xspace}

\newcommand{\twr}[1]{\textbf{{\color{cyan}\emph{T: #1}}}}
\definecolor{bottlegreen}{rgb}{0.0,0.42,0.31}

\newcommand{\Omit}[1]{}

\definecolor{mydarkblue}{rgb}{0,0.1,0.45}
\hypersetup{
   colorlinks=true,
   linkcolor=mydarkblue,
   citecolor=mydarkblue,
   filecolor=mydarkblue,
   urlcolor=mydarkblue}

\usepackage{ragged2e}

\usepackage{cleveref}
\usepackage[ruled]{algorithm2e}
\renewcommand{\@algocf@capt@plain}{above}
\makeatother
\usepackage{wrapfig}

\def\1{\bm{1}}

\def\sS{{\mathbb{S}}}

\def\sX{{\mathbb{X}}}
\def\sY{{\mathbb{Y}}}
\def\sZ{{\mathbb{Z}}}

\newcommand{\QED}{\mbox{$\Box$}}
\newcounter{my_counter}

\newcommand{\MyRoman}[1]%
        {\setcounter{my_counter}{#1}\roman{my_counter}}
\newcommand{\PMyRoman}[1]%
        {\rm (\MyRoman{#1})}
\newtheorem{MyTheorem}{Theorem}[section]

{\par\noindent{\bf Theorem~\ref{The:#1}\/}\begin{em}}%
{\end{em}$\QED$}
\newtheorem{Proposition}[MyTheorem]{Proposition}

{\par\noindent{\bf Proposition~\ref{Pro:#1}\/}\begin{em}}%
{\end{em}$\QED$}
\newtheorem{Lemma}[MyTheorem]{Lemma}

{\par\noindent{\bf Lemma~\ref{Lem:#1}\/}\begin{em}}%
{\end{em}$\QED$}
{\par\noindent\begin{rm}{\em Claim\/}:~#1}%
{\end{rm}}

\newtheorem{Corollary}[MyTheorem]{Corollary}

\newtheorem{Conjecture}[MyTheorem]{Conjecture}

\newtheorem{Observation}[MyTheorem]{Observation}

\newtheorem{Definition}[MyTheorem]{Definition}

\newtheorem{SDefinition}[MyTheorem]{Strawman Definition}

\newtheorem{Desideratum}[MyTheorem]{Desideratum}

\newtheorem{Example}[MyTheorem]{Example}

                        {$\QED$\end{quote}}

\usepackage{ragged2e}

\DeclareTextFontCommand{\emph}{\em}
\renewcommand{\sY}{\mathsf{Y}}
\renewcommand{\sX}{\mathsf{X}}
\renewcommand{\sZ}{\mathsf{Z}}
\newcommand{\Z}{{Z}}
\renewcommand{\sS}{\mathsf{S}^{\textrm{rhs}}}

\renewcommand{\Prog}{{Prog}}

\renewcommand{\S}{S}
\renewcommand{\paragraph}[1]{\vspace{5 pt} \noindent \textbf{#1}}
\newcommand{\symb}[1]{{\textit{#1}}}
\newcommand{\att}[1]{\textrm{#1}}

\usepackage{hyperref}

\title{Neural Program Generation Modulo Static Analysis}

\author{
Rohan Mukherjee \\
Rice University
\And
Yeming Wen \\
UT Austin 
\And 
Dipak Chaudhari\\
UT Austin 
\And 
Thomas W. Reps \\
University of Wisconsin 
\And 
Swarat Chaudhuri\\
UT Austin 
\And 
Chris Jermaine \\
Rice University 
}

\begin{document}
\maketitle 










\begin{abstract}

State-of-the-art neural models of source code tend to be evaluated on the generation of individual expressions and lines of code, and commonly fail on long-horizon tasks such as the generation of entire method bodies. 
We propose to address this deficiency using weak supervision from a static program analyzer. Our neurosymbolic method
allows a deep generative model to symbolically compute, using calls to a static-analysis tool, long-distance semantic relationships in the code that it has already generated. During training, the model observes these relationships and learns to generate programs conditioned on them. We apply our approach to the problem of generating entire Java methods given the remainder of the class that contains the method. Our experiments show that the approach substantially outperforms state-of-the-art transformers and a model that explicitly tries to learn program semantics on this task, both in terms of producing programs free of basic semantic errors and in terms of syntactically matching the ground truth.

\end{abstract}

\section{Introduction}\label{sec:intro}
Neural models of source code have received much attention in the recent past~\citep{YinN17,chen2018tree,bayou,ling2016latent,parisotto2016neuro,svyatkovskiy2020intellicode,gemmell2020relevance,lu2021codexglue}. 
However, these models have a basic weakness: while they frequently excel at 
generating individual expressions or lines of code, 
they do not do so well when tasked with synthesizing larger code blocks. For example, as we show later in this paper, state-of-the-art transformer models \citep{codex,gpt-neo,lu2021codexglue} 
can generate code with elementary semantic errors, such as uninitialized variables and type-incorrect expressions, when asked to generate \emph{method bodies}, as opposed to single lines. 
Even in terms of syntactic accuracy measures, the quality of the code that transformers produce on such ``long-horizon'' tasks can be far removed from the ground truth.

The root cause of these issues, we believe, is that current neural models of code treat programs as text rather than artifacts that are constructed following a \emph{semantics}. 
In principle, a model could learn semantics from syntax given enough data.
In practice, such learning is difficult for complex, general-purpose languages. 


  

In this paper, we propose to address this challenge through an alternative \emph{neurosymbolic} approach.
Our main observation is that symbolic methods---specifically, static program analysis---can extract deep semantic relationships between far-removed parts of a program. However, these relationships are not apparent at the level of syntax, and it is difficult for even large neural networks to learn them automatically. Driven by this observation, we use a static-analysis tool as a \emph{weak supervisor} for a deep model of code. During generation, our model invokes this static analyzer to compute a set of semantic facts about the code generated so far. The distribution over the model's next generation actions is conditioned on these facts. 

We concretely develop our approach by extending the classic formalism of {\em attribute grammars}~\cite{MST:Knuth68}. Attribute grammars are like context-free grammars but allow rules to carry symbolic {\em attributes} of the context in which a rule is fired.
In our model, called {\em Neurosymbolic Attribute Grammars} ({\nag}s), the context is an incomplete program, and rules are fired to replace a nonterminal (a stand-in for unknown code) in this program.
The attributes are semantic relationships (for example, symbol tables) computed using static analysis. The neural part of the model represents a probability distribution over the rules of the grammar conditioned on the attributes. During generation, the model repeatedly samples from this distribution while simultaneously computing the attributes of the generated code. 

We evaluate our approach in the task of generating the entire body of a Java method given the rest of the class in which the method occurs.
We consider a large corpus of curated Java
programs, over a large vocabulary of API methods and types.
Using this corpus, we train an \nag whose attributes, among other things, track the state of the symbol table and the types of arguments and return values of invoked methods at various points of a program, 
and whose neural component is a basic tree LSTM.
We compare this model against several recent models: fine-tuned versions of two \gptneo~\citep{gpt-neo} transformers and the \codegpt~\citep{lu2021codexglue} transformer, OpenAI's \codex system \citep{codex} (used in a zero-shot manner), 
and a GNN-based method for program encoding 
\citep{brockschmidt2018generative}. Some of these models are multiple orders of magnitude larger than our \nag model. 
Our experiments show that the {\nag} model reliably outperforms all of the baselines on our task, 
both in terms of producing programs free of semantic errors and in terms of matching the ground truth syntactically. 

In summary, this paper makes three contributions:
\begin{itemize}[leftmargin=0.6cm]
  \item
    We present a new approach to the generative modeling of source code that uses a static-analysis tool as a weak supervisor.
    
\item We embody this approach in the specific form of {\em neurosymbolic attribute grammars} ({\nag}s).
    
  \item We evaluate the {\nag} approach on the long-horizon task of generating entire Java method bodies, and show that it significantly outperforms several larger, state-of-the-art transformer models.
\end{itemize}

\section{Conditional Program Generation}\label{sec:overview}

\newcommand{\expec}{\mathbf{E}}


\begin{wrapfigure}{r}{0.4\columnwidth}
\vspace{-0.22in}
\begin{mdframed}
(a)

\begin{codesmall}
public class FileUtil{
  String err;
  public int read(File f){...}
  
  /* write lines to file */ 
  public void write(
    File f, String str){??}}
\end{codesmall}

(b) 

\begin{codesmall}
void write(File f, String str){
  try {
    FileWriter var_0;
    var_0 = new FileWriter(f);
    var_0.write(str);
  } catch(IOException var_0) {
    var_0.printStackTrace();
    System.out.println( ARG ); }
  return; }
\end{codesmall}
\end{mdframed}
\caption{(a) An instance of conditional program generation.  (b) A top completion of the \texttt{write} method, generated using an \nag. \texttt{ARG} stands for a string literal.}\label{fig:completion}
\vspace{-0.2in}
\end{wrapfigure}

We start by stating our problem, known as conditional program generation (CPG)~\citep{bayou}. We imagine a joint distribution $\mathcal{D}(X, Y)$, where $X$ ranges over \emph{specifications} of program-generation problems and $Y$ ranges over programs. The probability $\mathcal{D}(X = \sX, Y = \sY)$ is high when $\sY$ is a solution to $\sX$. Also, we consider a family of distributions $\mathcal{P}_{\theta}(Y | X = \sX)$, parameterized by $\theta$, that we might want to learn. \emph{Learning to conditionally generate} programs amounts to finding parameters $\theta$ that minimize the prediction error 
$
\expec_{(\sX, \sY) \sim \mathcal{D}}[\delta(\mathcal{P}_{\theta}(\sX | \sY), \sY)], 
$
where $\delta$ is a suitable distance function between programs.

Specifications and distances between programs can be defined in many ways. In our experiments, the goal is to generate Java method bodies. 
A specification is an \emph{evidence set} that contains information---e.g., method names, types of variables and methods---about the class in which the method lies. We define $\delta(\sY_1, \sY_2)$ to be a large number if $\sY_1$ or $\sY_2$ violates one of several language-level invariants (e.g., type-safety, initialization of variables before use) that we require programs to satisfy. When both programs satisfy the invariants, $\delta(\sY_1, \sY_2)$ measures the textual dissimilarity between the two programs. 


Note that CPG is a much more challenging task than the well-studied next-token-prediction task \citep{lu2021codexglue,brockschmidt2018generative}. The goal is to predict long sequences of tokens (e.g., an entire method body). Also, $\sX$ is a (possibly imprecise) specification of the code to generate, not just a sequence of tokens we are trying to complete by, say, choosing the correct method to call for a variable.

\paragraph{Example.} 
Fig.~\ref{fig:completion}-(a) illustrates the kind of task that we target. 
Here, we are given a class with a missing \texttt{write} method. 
The specification $\sX$ includes: (i) the class name \texttt{FileUtil}; (ii) the type \texttt{String} of the class variable 
\texttt{err}; (iii) information about 
complete methods within the class 
(including the methods' return types and formal-parameter types and names, and sequences of API calls made within such methods); (iv) information about the method with missing code (\texttt{write}), including its name, formal parameters, and JavaDoc comments for the method with missing code (e.g., ``write lines to file''). Our objective on this input is to generate automatically a non-buggy, natural completion of \texttt{write}, without any provided, partial implementation of the method. 



To understand the challenges in this task, consider a completion that starts by: (i) 
declaring a local variable \texttt{var\_0}; and (ii)
invoking the constructor for \texttt{FileWriter} and storing the result in \texttt{var\_0}.
A proper implementation of these two steps must ensure that \texttt{var\_0} is of type \texttt{FileWriter}. Also, the first argument to the constructor of \texttt{FileWriter} must be 
of type \texttt{File} (or a subtype of \texttt{File}). 
As we show in Sec.~\ref{sec:eval}, it is hard for state-of-the-art neural models 
to learn to satisfy these rules.

In contrast, in our 
approach, 
the generator has access to a set of 
semantic attributes computed via static analysis. 
These attributes include a \emph{symbol table} mapping in-scope variables to their types. 

Suppose that during training we are given the following line of code:
``\texttt{var\_0} = \texttt{new}~\texttt{FileWriter(f, true)}''.
Our model's symbol table includes the names \texttt{var\_0} and \texttt{f} and their types. 
The grammar is also able to compute the type of the first argument in the invoked constructor for \texttt{FileWriter}.
Consequently, the model can observe that the type of \texttt{f} is listed as \texttt{File} in the symbol table, and that \texttt{f} is the first argument to the  \texttt{FileWriter} constructor.
With a few observations like these, the model can learn that the first argument of ``\texttt{new} \texttt{FileWriter}'' tends to be of type \texttt{File} (or a subtype).
During generation, the model uses this knowledge, locating a variable of the correct type in the symbol table each time it constructs a \texttt{FileWriter}.


Fig.~\ref{fig:completion}-(b) shows 
a top completion of \texttt{write} generated by our \nag implementation. 
Note that all variables in this code are initialized before use, and that all operations are type-safe. 
Also, the name \texttt{var\_0} is reused between the \textbf{try} and the \textbf{catch} blocks.
Such reuse is possible because the symbol table carries information about the scopes to which different names belong. Finally, as we will see in Sec.~\ref{sec:eval}, the extra information provided by the static analyzer can also help with accuracy in terms of syntactic matches with the ground truth.





\section{Static Analysis with Attribute Grammars}\label{sec:attribute}
As mentioned in Sec.~\ref{sec:intro}, we develop our approach as an extension of the classic attribute grammar (AG) framework \cite{MST:Knuth68}. Now we give some background on static analysis using AGs. In the next section, we show how to use AGs to weakly supervise a neural program generator. 

\begin{figure}
\begin{tabular}{|l|l|}
\hline
(a) & (b) \\ 
{\fontsize{8pt}{8pt}\selectfont
$
\begin{array}{r@{\hspace{0.5ex}}c@{\hspace{2.0ex}}l}
  \symb{Stmt}  & : & \symb{Stmt}\textbf{;}~\symb{Stmt} \\
      & \mid &
  \symb{Expr}~\textbf{.}~\symb{Method}~\textbf{(}\symb{ArgList}\textbf{)} \\
        & \mid & \symb{DeclType}~\symb{Var}~\textbf{=}~\textbf{new}~\symb{NewType}~\textbf{(}\symb{ArgList}\textbf{)}  \medskip 
\end{array}
$
}
&
{\fontsize{8pt}{8pt}\selectfont
$
\begin{array}{r@{\hspace{0.5ex}}c@{\hspace{2.0ex}}l}
  \symb{Stmt} & \{  & \textit{SymTab}~\att{symTab} \downarrow; ~~\textit{SymTab}~\att{symTabOut} \uparrow; \}; 
\end{array}
$
} 
\\
\hline
(c) & (d) \\
\begin{minipage}{0.48\linewidth}
{\fontsize{8pt}{8pt}\selectfont
$$
\begin{array}{r@{\hspace{0.5ex}}c@{\hspace{2.0ex}}l}
  \symb{Stmt} & : &
  \symb{Stmt}\textbf{;}~\symb{Stmt}\textbf{;} \\
  & &  [~\symb{Stmt}\$1.\att{symTab}\downarrow~:=~\symb{Stmt}\$0.\att{symTab}\downarrow \\ 
                && ~\symb{Stmt}\$2.\att{symTab}\downarrow~:=~\symb{Stmt}\$1.\att{symTabOut}\uparrow \\ 
                && ~\symb{Stmt}\$0.\att{symTabOut}\uparrow~:=~\symb{Stmt}\$2.\att{symTabOut}\uparrow ] \medskip \\ 
  \symb{Stmt}  & : &
  \symb{Expr}~\textbf{.}~\symb{Method}~\textbf{(}\symb{ArgList}\textbf{)}~\\
  & & ~~[~\symb{Stmt}.\att{symTabOut}\uparrow~:=~\symb{Stmt}.\att{symTab}\downarrow \ldots~] \smallskip \\
        & \mid & \symb{DeclType}~\symb{Var}~\textbf{=}~\textbf{new}~\symb{NewType}~\textbf{(}\symb{ArgList}\textbf{)} \\
        & & ~~[  
           ~\symb{DeclType}.\att{symTab}\downarrow~:=~\symb{Stmt}.\att{symTab}\downarrow\\
            &&   
              ~~~~\symb{Var}.\att{symTab}\downarrow~:=~\symb{Stmt}.\att{symTab}\downarrow\\
            && ~~~~\symb{Stmt}.\att{symTabOut}\uparrow~:=~\symb{Stmt}.\att{symTab}\downarrow + \\
           &&   \qquad (\symb{Var}.\att{name}\uparrow~\mapsto~\symb{DeclType}.\att{type}\uparrow) \\
            &&   
              ~~~~\symb{ArgList}.\att{typeList}\downarrow~:=~\symb{NewType}.\att{typeList}\uparrow \\
           &&   
              ~~~~\symb{ArgList}.\att{symTab}\downarrow~:=~\symb{Stmt}.\att{symTab}\downarrow \\
           &&   
              ~~~~\symb{NewType}.\att{declType}\downarrow~:=~\symb{DeclType}.\att{type}\uparrow ]
\end{array}
$$ 
} 
\end{minipage}
& 
\begin{minipage}{0.35\linewidth}
{\centering 
\includegraphics[width=1.37\linewidth, trim={0 0 0 0.cm}, clip]{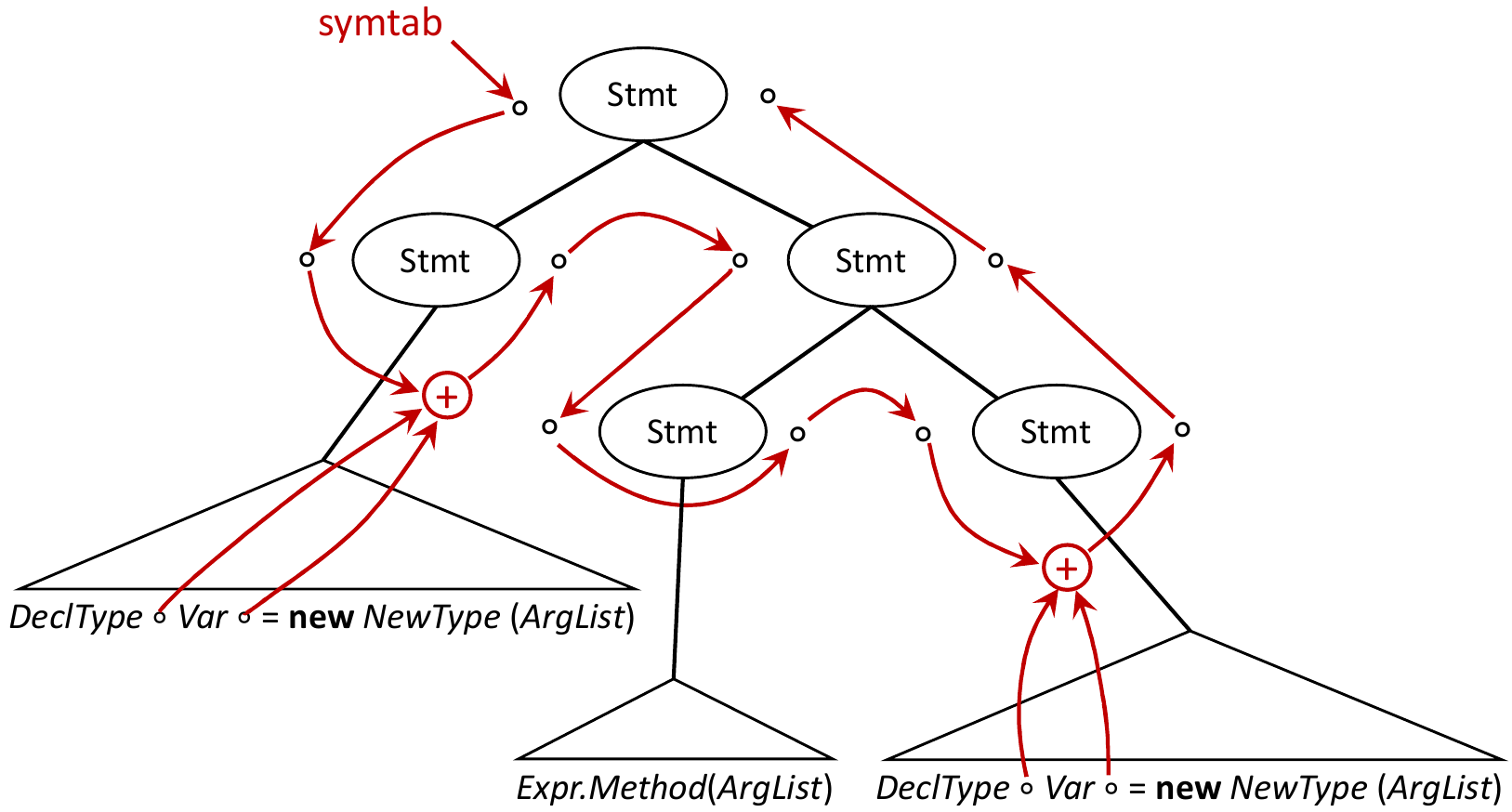}
} 
\end{minipage}\\
\hline 
\end{tabular}
\caption{(a) A basic context-free grammar.~~(b) Attributes of the $\symb{Stmt}$ nonterminal. ~~(c) Attribute equations for the productions (the parts of the equations denoted by ``\dots'' are elided).~~ (d) An attributed tree, illustrating left-to-right threading of attributes.}
\label{fig:grammar}
\vspace{-0.2in}
\end{figure}

An AG extends a traditional context-free grammar (CFG)~\cite{hopcroft2001introduction} by attaching a set of \emph{attributes} to each terminal or nonterminal symbol of the grammar and by using a set of \emph{attribute equations} to propagate attribute values through syntax trees. 
The attributes of a symbol $\S$ can be divided into
\emph{inherited} attributes and \emph{synthesized} attributes, which we suffix by $\downarrow$ and $\uparrow$, respectively.
Inherited attributes transfer information from parent to child, or from a node to itself.
Synthesized attributes transfer information from child to parent, from a node to a sibling, or from a node to itself. We assume that the terminal symbols of the grammar have no synthesized attributes and that the root symbol of the grammar has a special set of inherited attributes, known as the \emph{initial attributes}.

The \emph{output} attributes of a production $\S \rightarrow \S_1, \ldots, \S_k$ consist of the synthesized-attribute occurrences of the nonterminal $\S$, plus the inherited-attribute occurrences of all of the $\S_i$'s.
The \emph{input} attributes are the inherited-attribute occurrences of $\S$, plus the synthesized-attribute occurrences of the $\S_i$'s. The grammar's attribute equations relate the input and output attributes of a node in terms of the attributes of its parent, children, and left sibling in the syntax tree that the grammar generates. 

\paragraph{Example.} Consider the simple CFG in Fig.~\ref{fig:grammar}-(a). The nonterminal $\symb{Stmt}$ stands for \emph{program statements}.
The grammar says that a statement can either be a sequential composition of statements, a method call, or a variable declaration.
A natural AG extension of this CFG tracks symbol tables, which allow easy lookup of all variables in scope. 

Specifically, the grammar associates two symbol-table-valued attributes, $\att{symTab} \downarrow$ and $\att{symTabOut} \uparrow$, with $\symb{Stmt}$ (Fig.~\ref{fig:grammar}-(b)). 
The attributes are propagated following the equations in Fig.~\ref{fig:grammar}-(c). 
In these equations, 
  we distinguish between the three different occurrences of nonterminal ``$\symb{Stmt}$'' via the symbols ``$\symb{Stmt}\$0$,'' ``$\symb{Stmt}\$1$,'' and ``$\symb{Stmt}\$2$.'' where the numbers denote the leftmost occurrence, the
  next-to-leftmost occurrence, etc. In this case, the leftmost occurrence is the left-hand-side occurrence.

For concreteness, let us consider the attribute equations for the production for sequential composition in the grammar. Here, the inherited attribute of $\symb{Stmt}\$0$ gets passed ``down'' the syntax tree as an inherited attribute of $\symb{Stmt}\$1$. The synthesized attribute received at $\symb{Stmt}\$1$ is passed to $\symb{Stmt}\$2$ as an inherited attribute. 
More generally, the attribute equations define a left-to-right information flow through the syntax tree, as illustrated in Fig.~\ref{fig:grammar}-(d).

\section{Neurosymbolic Attribute Grammars} \label{sec:condproggen}

Now we introduce the model of neurosymbolic attribute grammars ({\nag}s). Our goal is to learn 
a distribution $\mathcal{P}(Y | \sX),$ where $Y$ is a random variable whose domain is all possible programs (concretely, Java method bodies) and $\sX$ is a specification of a program-generation problem (concretely, an \emph{evidence set} made up of useful information extracted symbolically from the method's context and then encoded using a neural network). 
Attributes containing the results of a symbolic, static analysis 
are available to the neural network implementing this distribution. This weak supervision allows the network to mimic more accurately the long-range dependencies present in real code-bases.

\newcommand{\bayou}{\textsc{Bayou}}

\paragraph{The Underlying Model.}~ The idea of weak supervision using a static analyzer could be developed on top of many different kinds of models. Here, we develop the idea on top of a model from \citet{bayou}.  This model uses a latent variable $\Z$ to represent the \emph{true user intent} behind the incomplete or ambiguous evidence set $Y$. We then have $\mathcal{P}(\sY|\sX) = \int_\sZ \mathcal{P}(\sZ|\sX) \mathcal{P}(\sY|\sZ) d\sZ. \label{eqn:cond}$
To define the distribution $\mathcal{P}(\Z|\sX)$, we assume that the evidence set has data of a fixed number of \emph{types}---e.g., method names, formal parameters, and Javadoc comments.

The $j^{\textit{th}}$ type of evidence has a neural encoder $f_j$. An individual piece of evidence $\sX$ is either encoded as a single vector or as a set of vectors with no particular ordering. For example, our implementation encodes Javadoc comments as vectors using LSTMs, and each member of a set of formal parameters using a basic feedforward network.
Let $\sX_{j,k}$ refer to the $k^{\textit{th}}$ instance of the $j^{\textit{th}}$ kind of evidence in $\sX$. 
Assume a Normal prior on $\Z$, and let
$\mathcal{P}(\sX | \sZ) = \prod\limits_{j,k} \mathcal{N} \left(f_{j}(\sX_{j,k}) \mid 
 		\sZ, \textbf{I} \sigma^2_j\right)$. 
 Assume that the encoding of each type of evidence is sampled from a Normal centered at $\sZ$.
 If $f$ is 1-1 and onto, 
we have \cite{bayou}:
\begin{equation} \nonumber
\mathcal{P}(\sZ | \sX) = \mathcal{N} \left(\sZ \mid \frac{\sum\limits_{j,k} \sigma^{-2}_j f_{j} (\sX_{j,k})}{1 + \sum\limits_j |\sX_{j}| \sigma^{-2}_j}, \frac{1}{1 + \sum\limits_j |\sX_{j}| \sigma^{-2}_j}\textbf{I}\right)
\end{equation}

\begin{wrapfigure}{rt}{0.52\textwidth}
\begin{minipage}{0.5\textwidth}
\begin{algorithm}[H]
\caption{Gen$(\S, A(S){\downarrow}, \mathsf{SymSoFar}, \sZ)$}
\smallskip 

\textbf{Input}: current symbol $\S$, inherited attributes $A(S){\downarrow}$, sequence of symbols so far $\mathsf{SymSoFar}$, latent encoding $\sZ$ \smallskip \\
\textbf{Modifies}: all symbols expanded are appended to $\mathsf{SymSoFar}$ \smallskip \\
\textbf{Returns}: $A(S){\uparrow}$, the synthesized attrs of $S$ \\

\vspace{5 pt}
\eIf{$S \textrm{ is a terminal symbol}$}{
  Append $(\S, \epsilon)$ to $\mathsf{SymSoFar}$ \\
  \Return $\emptyset$
}{
Choose a right-hand-side (RHS) sequence \\
$\textrm{  } \sS$ $\sim$ $\mathcal{P}(\S | \mathsf{SymSoFar}, A(S){\downarrow}, \sZ)$ \\
Append $(\S, \sS)$ to $\mathsf{SymSoFar}$ \\
$\mathsf{SynthSoFar} \gets \langle \rangle$ \\
\For{$\S' \in \sS \textrm{ in left-to-right order }$}{
        Compute $A(S'){\downarrow}$ from $A(S){\downarrow}$ and $\mathsf{SynthSoFar}$ \\
        $A(S'){\uparrow}$ $\leftarrow$ Gen$(\S', A(S'){\downarrow}, \mathsf{SymSoFar}, \sZ)$ \\
        Append $A(S'){\uparrow}$ to $\mathsf{SynthSoFar}$ 
     }
}
Compute $A(S){\uparrow}$ from $A(S){\downarrow}$ and $\mathsf{SynthSoFar}$ \\
\Return $A(S){\uparrow}$
\end{algorithm}
\end{minipage}
\end{wrapfigure}

Next, we define the distribution $\mathcal{P}(Y|\sZ)$. Consider a stochastic CFG which assumes (1) that a leftmost derivation is carried out, and (2) the probability distribution governing the expansion of a symbol in the grammar takes into account the sequence of all expansions so far, as well as an input value $\sZ$ upon which all expansions are conditioned.

This CFG consists of productions of the form
$\S ~:~ seq_1 \textrm{ }|\textrm{ } seq_2 \textrm{ }|\textrm{ } seq_3... \textrm{ }|\textrm{ } seq_n$.
Each symbol such as $\S$ corresponds to a categorical random variable with sample space $\Omega(\S) = \{seq_1, seq_2, ..., seq_n\}$. A trial over the symbol $\S$ randomly selects one of the RHS sequences for that symbol. If $\S$ is a terminal symbol, then $\Omega(\S) = \{\epsilon\}$, where $\epsilon$ is a special value that cannot be expanded. 
Subsequently, when a trial over $\S$ is performed and an RHS sequence from $\Omega(\S)$ is randomly selected, we will use the sans-serif $\sS$ to denote the identity of the RHS sequence observed.  

Now consider a depth-first, left-to-right algorithm for non-deterministically expanding rules in the grammar to generate a \emph{program} $\sY = \langle (\S_1, \sS_1), (\S_2, \sS_2), ...\rangle$; here, each $\S_i$ is a symbol encountered during the expansion, and each $\sS_i$ is the identity of the RHS chosen for that symbol.  Let $\S_1$ correspond to the symbol $\id{Start}$.
We perform a trial over $\S_1$ and select one of the RHS sequences from $\Omega(\S_1)$. 
Let the identity of the RHS sequence selected be $\sS_1$.
Note that $\sS_1$ is \emph{itself} a sequence of symbols.  
Choose the first symbol in the sequence $\sS_1$; call this symbol $\S_2$.
Perform a trial over $\S_2$, and let the identity of the RHS sequence chosen be $\sS_2$.
Choose the first symbol in $\sS_2$ (call it $\S_3$) and expand it the same way.  This recursive descent continues until a terminal symbol $\S_i$ is encountered, and the recursion unwinds.  If the recursion unwinds to symbol $\S_2$, for example, then we choose the second symbol in the sequence $\sS_1$, which we call $\S_{i + 1}$.  We perform a trial over $\S_{i + 1}$, and let the identity of the RHS sequence chosen be $\sS_{i + 1}$.  This sequence is recursively expanded.  
Once all of the symbols in the RHS associated with the $\id{Start}$ symbol $\S_1$ have been fully expanded, we have  a program.

This generative process defines a probability distribution $\mathcal{P} (Y | \sZ)$, where for a particular program $\sY$, the probability of observing $\sY$ is computed as
\begin{align}
\mathcal{P}(\sY | \sZ) = \prod_i & \mathcal{P}(\S_i = \sS_i |  \S_1 = \sS_1, ..., \S_{i - 1} = \sS_{i - 1}, \sZ). \label{eqn:conditionalgeneration}
\end{align}
We henceforth abbreviate the expression for the inner probability as $\mathcal{P}(\sS_i | \sS_1, ..., \sS_{i - 1}, \sZ)$.

\paragraph{Weak Supervision with Attributes.}~ Now assume that the grammar is an AG, so that each symbol $\S$ has an attribute set $A(\S)$. We use $A(\S){\uparrow}$ to denote the synthesized attributes of $\S$, and $A(\S){\downarrow}$ to denote the inherited attributes of $\S$.

An \nag extends the model so that the conditional distribution $\mathcal{P}(Y | \sZ)$ is defined as:
\begin{align}
\mathcal{P}(\sY | \sZ) = \prod_i \mathcal{P}(\sS_i | \langle \sS_{1}, \sS_{2}, ..., \sS_{i-1} \rangle,
A(\S_i){\downarrow}, 
\sZ). \nonumber
\end{align}
\noindent
That is, when a symbol $S_i$ is non-deterministically expanded, its value depends not just on the latent position $\sZ$ and the sequence of expansions thus far, but also on the values of
$S_i$'s inherited attributes, $A(\S_i){\downarrow}$.
In theory, a powerful enough learner with enough data could learn the importance of these sets of attribute values, without ever seeing them explicitly.
In that sense, they could be treated as latent variables to be learned.
However, the benefit of having a static analysis produce these values deterministically is that the author of a static analysis knows the semantic rules that must be followed by a program;
by presenting the data used to check whether those rules are followed directly to a learner, the process of learning to generate programs is made much easier.

Generation of a program under an \nag is described in Algorithm 1, where the distribution governing the expansion of symbol $S$ has access to attribute values $A(S){\downarrow}$.

\textbf{Designing an appropriate static analysis.} Intuitively, a program generated with the supervision of a static analyzer is likely to generate a semantically correct program because the static analysis provides key semantic clues during program generation.
In a conventional AG-based analyzer, the AG would be used to maintain data structures that can be used to \emph{validate}
that in a complete program, key relationships hold among the values of the production’s attributes.
Our goal is to generate programs, rather than validate them;
also, we want to \emph{guide} the learner rather than impose hard constraints.
However, constraints are a good mental model for designing a good \nag. 
That is, we generally expect the attribute equations used at important decision points during a neural generation process to be also 
helpful for validating key semantic properties of complete programs. 



\textbf{Example.} Now we show how to use the attribute grammar in Fig.~\ref{fig:grammar} in generating the body of the \texttt{write} method from Sec.\ \ref{sec:overview}. 
Let us assume that the grammar has a start nonterminal $\symb{Start}$ (not shown in Fig.~\ref{fig:grammar}) that appears in a single rule expanding it to the statement nonterminal $\symb{Stmt}$.
We start by extracting the context $\sX$ around the method, then use this information to sample $\sZ$ from $P(Z | \sX)$.
Next, a Java compiler processes the surrounding code and the method's formal parameters to form the attributes $A(\symb{Start}){\downarrow}$, which we assume to consist of a symbol table $\{~ \texttt{f} \mapsto \texttt{File}, \texttt{str} \mapsto \texttt{String} ~\}$.

To generate a program, we sample from the distribution $P(\symb{Start} | \langle \rangle, A(\symb{Start}){\downarrow}, \sZ)$.
First, $\symb{Start}$ is expanded to ``$\symb{Stmt}~\textbf{;} ~\symb{Stmt}$''.
When expanding the first $\symb{Stmt}$, the \nag needs to choose between a method invocation and a variable declaration. 
Because the \nag is ``aware'' that this step is to expand the first line of the method---the list of RHS values chosen so far is empty---we would expect it to declare a variable.
This choice gives us the RHS  ``$\symb{DeclType}~\symb{Var}~\textbf{=}~\textbf{new}~\symb{NewType}~\textbf{(}\symb{ArgList}\textbf{)}$''. Expanding $\symb{DeclType}$, the \nag samples a Java type from the distribution 
\begin{align}
P(&\symb{DeclType} | \langle\textrm{``}\symb{Stmt}\textbf{;}~\symb{Stmt}\textrm{''}, \textrm{``}\symb{DeclType}~\symb{Var}~\textbf{=}~ \textbf{new}~\symb{NewType}~\textbf{(}\symb{ArgList}\textbf{)}\textrm{''} \rangle, A(\symb{DeclType}){\downarrow}, \sZ). \nonumber
\end{align}
From the rules for expanding the nonterminal $\symb{DeclType}$ in Fig.~\ref{fig:grammar}, we see that 
the \nag can choose any Java type as the declared type of the variable. At this point, the \nag 
is aware that the goal is to create a method called \texttt{write} (this is encoded in $\sZ$) and that it is choosing a type to be declared on the first line of the method. 
It also has access to the symbol table that is maintained as part of $A(\symb{DeclType}){\downarrow}$.  Thus, the \nag may decide to expand the symbol $\symb{DeclType}$ to \texttt{FileWriter}. This type is then passed upward via the synthesized attribute $\symb{DeclType}.\att{type}{\uparrow}$.  

Next, the grammar must expand the $\symb{Var}$ rule and pick a variable name to declare.
This choice is returned via the synthesized attribute  $\symb{Var}.\att{name}{\uparrow}$. Now it is time to expand $\symb{NewType}$.
The attributes make this easy: when sampling from $P(\symb{NewType} | ...)$, the \nag has access to $\symb{NewType}.\att{type}{\downarrow}$, which takes the value \texttt{FileWriter}.
A synthesizer may err by choosing a type that is not compatible with \texttt{FileWriter}.  However, we may expect that during training, every time that $\symb{NewType}$ was expanded and the declared type was \texttt{FileWriter}, the type chosen was either \texttt{FileWriter} or some subclass of \texttt{FileWriter}. Hence the \nag is unlikely to make an error.  

Assume that the \nag chooses \texttt{FileWriter}. It must now expand $\symb{ArgList}$.
Again, the \nag has the advantage of having access to $\symb{ArgList}.\att{typeList}{\downarrow}$ (an explicit representation of the types required by the constructor being called) and, most importantly, $\symb{ArgList}.\att{symTab}{\downarrow}$ 
(an explicit list of the variables in scope, as well as their types).  
At this point, it is easy for the \nag to match the required type of the first argument to the constructor (\texttt{File}) with an appropriate variable in the symbol table (\texttt{f}).  

Now that the declaration of \texttt{var\_0} has been fully expanded, the \nag updates the symbol table with a binding for the newly-declared variable \texttt{var\_0}, and the attribute $\symb{Stmt}.\att{symTab}{\uparrow}$ takes the value 
$\{ \texttt{f} \mapsto \texttt{File}, \texttt{str} \mapsto \texttt{String}, \texttt{var\_0} \mapsto \texttt{FileWriter} \}$.
When the second occurrence of $\symb{Stmt}$ is expanded, the symbol table is passed down via the inherited attribute $\symb{Stmt}\$1.\att{symTab}\downarrow$.
All of the information available---the latent variable $\sZ$ encoding the contextual information (including the name of the method ``\texttt{write}'' being generated), and the symbol table containing a \texttt{FileWriter} and a \texttt{String})---helps the \nag to deduce correctly that this $\symb{Stmt}$ symbol should be expanded into an invocation of a \texttt{write} method.
Also, the presence of the symbol table makes it easy for the \nag to correctly attach the \texttt{write} method call to the variable \texttt{var\_0} and to use \texttt{str} as the argument.  



\section{Evaluation}\label{sec:eval}
Our experimental hypothesis is that neural networks find it difficult to learn the intricate rules 
that govern the generation of code by only looking at the syntax of example programs.
These issues become especially visible when the units of code to be generated are large, for example, entire method bodies.
In contrast, an \nag can use its static analysis to compute long-distance dependencies between program variables and statements ``for free.''
Because of this extra power, {\nag}s can outperform much larger neural models at generating accurate and semantically correct code. 


\subsection{Experimental Setup}

\textbf{Data}.  To test our hypothesis, we used a curated, deduplicated set of Java source-code files \cite{bayou}.
For each class and each method, we used the remainder of the class as evidence or context, and the method body was used to produce training or test data.
We used 1.57 M method bodies for training. The grammar used had ten terminals corresponding to formal parameters, ten for class variables, and ten for methods local to the class. None of the Java classes in the corpus needed more than ten of each of these terminals; when generating training data, each declared Java variable or method was randomly mapped to one of the appropriate terminals.  Approximately 8,000 types and 27,000 method calls from the Java JDK also appeared as terminals in the grammar. 

\textbf{{\nag} Implementation.} 
We implemented an \nag for our subset of Java. Here, attributes are used to keep track of
the state of the symbol table,
the expected return type of each method,
expected types of actual parameters,
variable initialization, whether the variable has been used,
and whether the method has a return statement.
The symbol table contains entries for all formal parameters, class variables, and internal methods within the class. 

The neural part of our model has 63 M parameters.
To expose the attributes to the neural part of the model, 
we implement a depth-first search over a program's abstract syntax tree (AST) to extract node information. The attributes are then encoded in a standard way --- for example, the symbol table is represented as matrix (rows correspond to types, columns to variables, the value 1 is present if the corresponding type/variable pair is in scope).  The distribution $\mathcal{P}(\sS_i | \langle \sS_{1}, \sS_{2}, ..., \sS_{i-1} \rangle,
A(\S_i){\downarrow}, 
\sZ)$ is implemented as a set of LSTMs that decode the sequence of symbols, as well as the encoded $A(\S_i){\downarrow}$ and $\sZ$, into a distribution over $\sS_i$. 
We trained our framework on top of Tensorflow \cite{tensorflow2015-whitepaper}. Using one GPU, the \nag training time is around 72 hours. See Appendix~\ref{sec:implementation} for more details.\footnote{Our implementation is available at \url{https://github.com/rohanmukh/nsg}.}

\textbf{Baselines.} 
We consider three categories of baselines. 
The first consists of large pretrained transformers. 
Specifically, we consider two variants of the \gptneo~\citep{gpt-neo} model with 125 M and 1.3 B parameters. Both models are pre-trained on the Pile dataset~\citep{Gao2021ThePA}, which consists of an 800 GB English-text corpus and open-source code repositories. On the APPS dataset~\citep{hendrycksapps2021}, they perform well compared to OpenAI's 12-B-parameter, GPT-3-like \codex model~\citep{codex}. We also compare against \codegpt~\citep{lu2021codexglue} which is a GPT-2-like model with 125 million parameters. This model was pre-trained on Python and Java corpora from the CodeSearchNet dataset, which consists of 1.1 M Python functions and 1.6 M Java methods.  We fine-tune all of these pretrained models on our Java dataset, using the token-level code-completion task provided by CodeXGLUE 
~\citep{lu2021codexglue}. Finally, we also offer a comparison against the \codex model~\citep{codex}. 
Because we did not have access to the model's pretrained weights, this model is only used in a zero-shot fashion (no fine-tuning on our Java dataset). It should be noted here that the transformer baselines work on the entire Java language, whereas our \nag framework works on a sub-part of Java which is supported in our grammar definition.

The second category comprises an ablation, called 
a 
``conditional neural grammar'' (CNG), that is identical to our \nag model but is trained without any of the attribute information. In other words, the CNG model is trained only on the program syntax.  
The third category includes GNN2NAG~\citep{brockschmidt2018generative}, a graph-neural-network-based method that uses an attribute grammar but \emph{learns} the attributes from data rather than computing them symbolically. See Appendix~\ref{sec:implementation} for more details on the baselines.

\textbf{Test Scenario.} Our test scenario is as follows. Given a Java class, we remove the entire body of a randomly selected method. We then use the remaining potion of the class along with the method header as context information that is  then fed to the model as input.
We run our \nag model and the baselines to regenerate this method body conditioned on the resulting context. We report the accuracy of the prediction based on static-semantic checks and fidelity measures. 

\begin{table*}[t] 
	\centering
	\vspace{-8 pt}
	\caption{ Percent of Static Checks Passed}
	\small
	\resizebox{\columnwidth}{!}{%
	\begin{tabular}{|l|c|c|c|c|c|c|c|c|}
		\hline
		\textit & GPTNeo125M & GPTNeo1.3B & \codex  & \codegpt & GNN2NAG & CNG & \nag  \\
		\hline
		\hline
		No undeclared variable access    & 89.87\% &  90.36\% & 88.62\% & 90.94\% & 47.44\% & {19.78\%} & \textbf{99.82\%}\\
		\hline
		Valid formal parameter access    & NA & NA & NA & NA &   25.78\% & {11.03\%} & \textbf{99.55\%}\\
		\hline
		Valid class variable access    &  NA & NA & NA & NA  & 15.40\% & {12.75\%} & \textbf{99.53\%} \\
		\hline
		No uninitialized objects    &  93.90\% & 91.73\%   & 90.82\% & 94.37\%  & 21.20\% &{21.56\%} & \textbf{99.01\%} \\
		\hline
        \hline
        No variable access error    & 90.36\% & 90.51\% & 88.86\% & 91.32\%  & 28.92\% & 17.92\% & \textbf{99.69\%}  \\
        \hline
        \hline
        Object-method compatibility & \textbf{98.36\%} & 98.09\% & 98.35\% & 97.84\%  & 21.43\% & 12.23\% & 97.53\%  \\
        \hline
        Return type at call site     & 97.38\% & 98.01\% & \textbf{98.53\%} & 97.83\%  & 23.86\% & 16.40\% & {98.01\%} \\
        \hline
        Actual parameter type       & 87.03\% & 86.36\% & 92.28\% & 88.71\%  & 9.27\% & 16.09\% & \textbf{97.96\%} \\
        \hline
        Return statement type         & 84.05\% & 85.09\% & 88.13\% & 85.23\%  & 12.34\% & 9.51\%  & \textbf{90.97\%} \\
        \hline
        \hline
        No type errors      & 87.25\% & 88.13\% & 91.42\% & 88.10\%  & 16.31\% & 13.56\%  & \textbf{97.08\%} \\
        \hline
        \hline
        Return statement exists                 & 99.61\% & 99.80\% & 98.44\% & 99.57\%  & 94.02\% & \textbf{99.92\%}  & {97.10\%} \\
        \hline
        No unused variables                 & 96.42\% & 96.46\% & 96.82\% & \textbf{97.64\%} & 20.95\% & 24.29\% & 93.84\%  \\
        \hline
        \hline
         Percentage of parsing           & 98.18\% & 98.13\% & 96.41\% & 97.08\% & \textbf{100.0}\% & \textbf{100.0}\% & \textbf{100.0}\% \\
         \hline
         Pass all checks  &  65.26\% & 64.88\% & 47.49\% & 67.73\% & 17.34\% & 12.87\% & \textbf{86.41\%} \\
         \hline
	\end{tabular}
	}
	\label{table:result_semantic_check_category}
\end{table*}

\subsection{Results} \label{sec:quant_results}

\textbf{Static Checks.}
For each generated method body, we check the following properties:
\emph{(1) No undeclared variable access:} Are all the variables used in a program declared (within an enclosing scope) before they are used?
\emph{(2) Valid formal parameter access:} Are formal parameters that are used in the method body present in the method declaration?  \emph{(3) Valid class-variable access:} Are the class variables that are used in the method body present in the class declaration? 
\emph{(4) No uninitialized objects:} Do variables have a non-null value when they are used? 
\emph{(5) No variable access errors:} Are checks (1)-(4) all satisfied?
\emph{(6) Object-method compatibility:} Are methods called on objects of a given class 
actually available within that class?
\emph{(7) Return type at the call site:} Is the assignment of the return value type-correct with respect to the return type of the called method? 
\emph{(8) Actual-parameter type:} Are the actual-parameter types in an API call consistent with the corresponding formal-parameter types?
\emph{(9) Return-statement type:} Is the type of the expression in a return statement consistent with the method's declared return type?
\emph{(10) No type errors:} Are checks (6)-(10) all satisfied? 
\emph{(11) Return statement exists:} Does the method body have a return statement somewhere?
 \emph{(12) No unused variables:} Are all variables declared in the method body used in the method?
 \emph{(13) Percentage of parsing:} Can the generated method be parsed by a standard Java parser?
 \emph{(14) Pass all checks:} Are checks (1)-(13) all satisfied?

Note that (2) and (3) are not meaningful metrics for approaches, such as our transformer baselines, that do not use a grammar to generate code. This is because in these models, when a variable token is generated, there is no way to tell what category of variable (class variable, formal parameter, etc.) it is meant to be. These metrics are meaningful for the \nag, CNG, and GNN2NAG models, which use a Java parser capable of partitioning variable names into different categories. 

The results of our comparisons appear in Table \ref{table:result_semantic_check_category}. 
These scores are interpreted as follows.
Suppose that a generated program uses five variables, of which four are declared correctly in the proper scope.
This situation is scored as 80\% correct on the "No undeclared-variable access" criterion. 
We report the average success rate over each of these properties over all the generated programs in our test suite.


\begin{table*}[t]
	\centering
	\caption{ Average Fidelity of Generated Method Bodies} 
	\small
	\resizebox{\columnwidth}{!}{%
	\begin{tabular}{|l|c|c|c|c|c|c|c|c|}
	    \hline
		\textit & GPTNeo125M & GPTNeo1.3B & \codex & \codegpt & GNN2NAG & CNG  & \nag \\
		\hline
		\hline
		Set of API Calls     &  32\% & 37\% & 36\%  & 36\% & 3\% & 22\% & \textbf{53\%} \\
		\hline
		Sequences of API Calls  &  17\% & 20\% & 16\% & 19\% & 0.3\% & 18\%  &\textbf{42\%}  \\
		\hline
        Sequences of Program Paths     & 12\% & 15\% & 10\% & 14\% & 0\% & 17\% &\textbf{39\%} \\
        \hline
        AST Exact Match                &  12\% & 15\% & 10\% & 14\% & 0\% & 6\% & \textbf{26\%} \\
        \hline
	\end{tabular}
    }
    \vspace{-0.3cm}
	\label{table:max_topk_jaccard}
\end{table*}


\textbf{Whole-Method Fidelity.} We also check the \emph{fidelity} of the generated code to the reference code.
One possibility here is to use a standard metric for text generation, such as the BLEU score.  However, this is problematic.
As the BLEU score is not invariant to variable renamings, a nonsensical program that uses commonplace variable names can get an artificially high BLEU score. Also, programs are structured objects in which some tokens indicate control flow and some indicate data flow. The BLEU score does not take this structure into account. See Appendix~\ref{appendix:bleu} for a concrete example of these issues.

\begin{wraptable}{l}{78mm}
    \centering
    \begin{tabular}{|p{1.2cm}|p{5.55cm}|}
    \hline 
   \scriptsize \textbf{Query} & 
   \vspace{-10 pt} \begin{codefootnote}
public class FileUtils{
 FileReader field_7;
 BufferedReader field_5;
   
 /** read line from file */
 public String reader () {}}
\end{codefootnote} \\
        \hline
        \scriptsize \textbf{\nag} & 
     \vspace{-10 pt} \begin{codefootnote}
public String reader(){
 java.lang.String var_9;
 try {var_9=field_5.readLine();
 } catch (IOException var_8) {
  var_8.printStackTrace(); }
 return var_9; }
\end{codefootnote} \\
        \hline
        \scriptsize \textbf{CodeGPT} &
        \vspace{-10 pt}\begin{codefootnote}
public String reader(){ 
   StringBuffer buffer=
     new StringBuffer(); 
   buffer.append("\n"); 
   return buffer.toString(); 
 }
\end{codefootnote} \\
        \hline
        \scriptsize \textbf{\codex} & \vspace{-10 pt}
       \begin{codefootnote}
public String reader() 
 throws IOException{ 
  field_5= new BufferedReader(
   new FileReader(field_7));
  return field_5.readLine(); }
\end{codefootnote} \\
        \hline
        \scriptsize \textbf{GptNeo1.3B} & \vspace{-10 pt}
       \begin{codefootnote}
public String reader() 
 throws IOException { 
  try { 
   field_7=new FileReader(this.file);
   field_5=new BufferedReader(field_7);
   String line; 
   while (field_5.readLine()) {
    System.out.println(line); }
    return line;
  } catch (FileNotFoundException e)
    {e.printStackTrace(); }
  return null; }
\end{codefootnote}


  \\
        \hline
    \end{tabular}
        \vspace{-5 pt}
    \caption{Reading from a file: Outputs for the \nag and transformer baselines.}
  \vspace{-10 pt}
  \label{table:completionGPT}
  \vspace{-2 pt}
\end{wraptable}

Instead, we consider four fidelity metrics:
(1) \emph{Set of API Calls}: Extract the set of API calls from the generated and reference codes, and compute the Jaccard similarity between the sets.
(2) \emph{Sequences of API Calls}: Generate the set of all possible API call sequences possible along code paths, and compute the Jaccard similarity between the sets for the generated and reference code.
(3) \emph{Sequences of Program Paths}: Generate the set of all possible paths from root to leaf in the AST, then compute the Jaccard similarity between the sets (two paths are equal if all elements 
except for object references match).
(4) \emph{AST Exact Match}: Exact AST match (except for object references), scored as 0 or 1.
We compute the highest value for each metric across the ten bodies generated, and average the highest across all test programs. Results for these measures are presented in Table~\ref{table:max_topk_jaccard}.

\textbf{Summary of results.}  We find that in most cases, the \nag had a higher incidence of passing the various static checks compared to the baselines.  This is perhaps not surprising, given that the \nag has access to the result of the static analysis via the attribute grammar. More intriguing is the much higher accuracy of the \nag for the fidelity results. Pre-trained language models and GNN2NAG are designed for next-token-prediction tasks (we give some results on these tasks in Appendix~\ref{sec:appendix_nexttoken}). However, in our CPG task, no tokens are available from the method body to be generated. In particular, language models must treat the surrounding code and method header as input from which to generate the entire method body, 
and this proves difficult. 
The \nag, on the other hand, uses static analysis to symbolically extract this context, which is explicitly given to the neural network in the form of the class variables and methods that are available to be called (in $A(S){\downarrow}$), and in the class name, encoded comments, variable names, and so on (in $\sZ$). 


\textbf{Transformers vs. {\nag}s}:  As a complement to our quantitative evaluation, we manually examined the outputs of our model and the baselines on a set of hand-written tasks for qualitative evaluation. The transformers produced impressively human-like code on several of these tasks. However, in quite a few cases, they
produced incorrect programs that a trained human programmer would be unlikely to write. Also, the transformers were biased towards producing short programs, which often led them to produce uninteresting outputs.


Table~\ref{table:completionGPT} illustrates some of the failure modes of the transformer baselines. Here, we consider the task of reading a string from a file utility class. The top result for our \nag model declares a \textit{String} variable to read from the already existing field while also correctly catching an \textit{IOException}. The \codegpt output in this case is unrelated to the context. \codex initiates a FileReader object by invoking an argument which is of type \textit{FileReader} itself, thereby causing a type mismatch. The code from \gptneo accesses a \textit{file} instance variable that does not exist and also returns a blank line from the method. 
A few other examples of \nag and transformer outputs appear in  Appendix~\ref{sec:additional_examples}. 




\section{Related Work}\label{sec:relwork}
\textbf{Non-Neural Models of Code.} 
Many non-neural models of code have been proposed over the years~\citep{ICML:MT14,raychev2014code,nguyenFSE13,idiomsFSE14,nguyenICSE15,bielik2016phog}.
A few of these models condition generation on symbolic information from the context. 
Specifically, \citet{bielik2016phog} use programmatically represented functions to gather information about the context in which productions for program generation are fired, then utilize this information to impose a distribution on rules. 
\citet{ICML:MT14} generate programs using a model that encodes a production's context using a set of  ``traversal variables.'' However, the absence of neural representations in these models puts a ceiling on their performance. 


\textbf{Deep Models of Code.}
There is, by now, a substantial literature on deep models trained on program syntax. Early work on this topic 
represented programs as sequences~\cite{raychev2014code} or trees~\cite{bayou,YinN17,chen2018tree}, and learned using classic neural models, such as RNNs, 
as well as specialized  
architectures~\cite{ling2016latent,parisotto2016neuro,alon2019code2vec}.
The recent trend is to use transformers~\cite{svyatkovskiy2020intellicode,gemmell2020relevance,feng2020codebert,lu2021codexglue}. 
Some of these models --- for example, \codegpt~\citep{lu2021codexglue} --- are trained purely on code corpora (spanning a variety of languages, including Java). Other models, such as \textsc{CodeBert}~\citep{feng2020codebert}, \gptneo~\citep{gpt-neo}, and \codex~\citep{codex}, are trained on both natural language and code. In all of these cases, programs are generated 
without any explicit knowledge of program syntax or semantics.


The GNN2NAG model by \citet{brockschmidt2018generative} also uses an attribute grammar to direct the generation of programs. However, unlike our method, this model 
use a graph neural net to \emph{learn} attributes of code. Our experiments show the benefits of our weak supervision approach over this. 

Also related is work by 
\citet{dai2018syntax}, who extend grammar variational autoencoders~\citep{kusner2017grammar} with hard constraints represented as attribute grammars. In that work, attribute constraints are propagated top-down, and every generated artifact is required to satisfy the top-level constraint. This strategy comes with challenges; as is well-known in the program-synthesis literature~\cite{synquid}, top-down constraint propagation can lead to unsatisfiability, and require rejection of generated samples, for grammars above a certain level of complexity. 
We sidestep this issue by using attribute grammars as a form of weak supervision, rather than as a means to enforce hard constraints.




\paragraph{Neurally Directed Program Synthesis.} 
Many recent papers study the problem of \emph{neurally directed program synthesis}~\cite{balog2016deepcoder,devlin2017robustfill,si2019learning,chen2020program,near,OdenaS20}. Here, neural networks, and sometimes program analysis, are used to guide a combinatorial search over programs. Because such search is expensive, these methods are typically limited to constrained domain-specific languages.
In contrast, our approach does not aim for a complete search over programs at generation time (our decoder does perform a beam search, but the width of this beam is limited). 
Instead, we embody our program generator as a neural network that sees program-analysis-derived facts as part of its data. 
This design choice makes our method more scalable 
and allows it to handle 
generation in a general-purpose language.

\vspace{-0.2cm}
\section{Conclusion}\label{sec:conc}
\vspace{-0.1cm}
We have presented a framework for deep generation of source code in which the training procedure is weakly supervised by a static analyzer, in particular, an attribute grammar.
We have shown that our implementation of this approach outperforms several larger, state-of-the-art transformers both in semantic properties and fidelity of generated method bodies.


A lesson of this work is that while modern transformers excel at writing superficially human-like code, they still lack the ability to learn the intricate semantics of general-purpose languages. 
At the same time, the semantics of code can be defined rigorously and partially extracted 
 ``for free'' using program analysis. This extracted semantics can be used to aid neural models with the concepts that they struggle with during program generation. While we have used this idea to extend a tree LSTM, we could have implemented it on top of a modern transformer as well. We hope that future work will pursue such implementations.

Our work demonstrates an alternative use for formal language semantics, compared to how semantics are typically used in program synthesis research. Historically, semantics have been used to direct generation-time combinatorial searches over programs. 
However, scalability has been a challenge with such approaches. 
Our work points to an alternative use of semantics: rather than using semantic program analyses to direct a search over programs, one could use them to \emph{annotate} programs at training and test time and leave the search to a modern neural network. We believe that such a strategy has the potential to vastly extend the capabilities of algorithms for program synthesis.


\textbf{Acknowledgments.} This work was funded by
NSF grants 1918651 and 2033851;
US Air Force and DARPA contract FA8750-20-C-0002;
and ONR grants N00014-17-1-2889, N00014-19-1-2318, and N00014-20-1-2115. We thank Matt Amodio for his help with our initial exploration of the problem studied here, and Martin Rinard and the anonymous referees for their valuable comments.


\bibliography{bibliography}
\bibliographystyle{icml2021}

\clearpage
\newpage

\appendix

\section{Additional Synthesis Examples} \label{sec:additional_examples}

Some additional program-generation examples are shown in Tables \ref{table:result_qual_reader}, \ref{table:result_qual_supp1}, and \ref{table:result_qual_supp2}.

\begin{table}
    \centering

    \begin{tabular}{|p{1.35cm}|p{6cm}|p{6cm}|}
        \hline
    & (a) Removing from a list & (b) Adding to a list   \\ 
        \hline
   \scriptsize \textbf{Query}  &
        \vspace{-10 pt}
        \begin{codefootnote}
public class myClass{

  /** remove items from a list */
  public void 
     remove(List<String> fp_2){
     }
}
\end{codefootnote}
 
& \vspace{-10 pt} 
    \begin{codefootnote}
public class myClass{

 /** add item to list */
 public void addItem (
  List<String> a, String b){
  }}
\end{codefootnote}
        \\
        \hline
    \scriptsize \textbf{\nag} &
        \vspace{-10 pt}
        \begin{codefootnote}
void remove(java.util.List
        <java.lang.Object> fp_2){
  java.util.Iterator var_8;
  var_8 = fp_2.iterator();
  while (var_8.hasNext()){
    java.lang.Object var_3;
    var_3 = new java.lang.Object(
      (java.lang.String) ARG, 
      (javax.swing.Icon) ARG);
    var_3 = var_8.next();
    var_8.remove();
}
return; 
}
\end{codefootnote}
 
& \vspace{-10 pt}
    \begin{codefootnote}
public void addItem (
     List<String> fp_9, 
     String fp_1) {
     if (fp_9.contains 
            ((String) ARG)) {
         fp_9.add(fp_1);
         }
    return; 
 }
\end{codefootnote}
        \\
        \hline
    \scriptsize \textbf{\codegpt } &
        \vspace{-10 pt}
        \begin{codefootnote}
public void remove(List<String> fp_2){ 
  items.remove(fp_2); }
\end{codefootnote}

&  \vspace{-10 pt}
\begin{codefootnote}
public void addItem (
        List<String> a, String b) 
 {
   items.add(a); 
 }
\end{codefootnote}  \\
        \hline
    \scriptsize \textbf{ \codex } &
        \vspace{-10 pt}
        \begin{codefootnote}
public void remove(List<String> fp_2){ 
  fp_2.remove(0); }
\end{codefootnote}
& \vspace{-10 pt}
\begin{codefootnote}
public void addItem (
 List<String> a, String b) {
 a.add(b); }
\end{codefootnote} \\
    \hline
    \scriptsize \textbf{ GptNeo1.3B } &
        \vspace{-10 pt}
        \begin{codefootnote}
public void remove(List<String> fp_2){ 
  fp_2.remove(); }
\end{codefootnote}
& \vspace{-10 pt}
\begin{codefootnote}
public void addItem (
 List<String> a, String b) {
    List <String> temp = 
           new ArrayList<String>();
   for (int i=0; i<a.size(); i++){
        temp.add(a.get(i)); 
    }
  a.add(b); 
 }
\end{codefootnote}  \\
\hline
    \end{tabular}

        \vspace{-5 pt}
    \caption{Example synthesis outputs: (a) Removing from a list and (b) Adding to a list.}
  \vspace{-5 pt}
        \label{table:result_qual_reader}
\end{table}



\begin{table}
    \centering
    \begin{tabular}{|p{1.35cm}|p{6cm}|p{6cm}|}
        \hline
    & (a) Writing to a file & (b) Creating a Swing JButton   \\ 
        \hline
    \scriptsize \textbf{ Query } &
        \vspace{-10 pt}
        \begin{codefootnote}
public class FileUtil{

  String err;
  public int read(File f){...}
  
  /* write lines to file */ 
  public void write(
    File f, String str){}}
\end{codefootnote} 
& \vspace{-10 pt} 
\begin{codefootnote}
public class myGUI{

   public javax.swing.JFrame 
         createFrame(final String title)
       {... }

   public void addActionListener
             (javax.swing.JButton title)
       {... }
       
   /**
   create a button
   */
   public 
    javax.swing.JButton createButton(){ }
}
\end{codefootnote}
 \\
        \hline
    \scriptsize \textbf{ \nag } &
        \vspace{-10 pt}
        \begin{codefootnote}
void write(File f, String str){
  try {
    FileWriter var_0;
    var_0 = new FileWriter(f);
    var_0.write(str);
  } catch(IOException var_0) {
    var_0.printStackTrace();
    System.out.println( ARG ); }
  return; }
\end{codefootnote}  
& \vspace{-10 pt}
    \begin{codefootnote}
javax.swing.JButton createButton(){
  javax.swing.ImageIcon var_9;
  var_9 = new javax.swing.ImageIcon.
                             ImageIcon(
              (java.lang.String) ARG,
              (java.lang.String) ARG
          );
  javax.swing.JButton var_7;
  var_7 = new javax.swing.JButton.JButton(
        java.lang.String: ARG,
        javax.swing.Icon: var_9);
  var_7.setVerticalTextPosition(
                (int) ARG);
  var_7.setHorizontalTextPosition(
               (int) ARG);
  return var_7;
}
\end{codefootnote}
 \\
    \hline
    \scriptsize \textbf{ \codegpt } &
        \vspace{-10 pt}
        \begin{codefootnote}
void write (File a, String b){
System.out.println(a);
}
\end{codefootnote} 
& \vspace{-10 pt}
  \begin{codefootnote}
public JButton createButton(){ 
    return new JButton(this);
}
\end{codefootnote}
 \\ 
        \hline
    \scriptsize \textbf{ CodeX } &
        \vspace{-10 pt}
        \begin{codefootnote}
public void write(File f, String str){
  Writer writer = null;                              
  try {                              
    writer = new BufferedWriter(
      new FileWriter(f));                              
    writer.write(str);                              
    writer.close();                              
  } catch (IOException e) {                              
    err = e.printStackTrace();}
}
\end{codefootnote} 
& \vspace{-10 pt}
  \begin{codefootnote}
public JButton createButton(){ 
    final javax.swing.JButton b = 
      new javax.swing.JButton();
    return b;
}
\end{codefootnote} \\ 
        \hline
    \scriptsize \textbf{ GptNeo1.3B } &
        \vspace{-10 pt}
        \begin{codefootnote}
public void write(File f, String str)
  { try { f.write(str); } 
  catch ( IOException e) { 
    e.printStackTrace(); }
}
\end{codefootnote} 
& \vspace{-10 pt}
  \begin{codefootnote}
public JButton createButton(){ 
    javax.swing.JButton button = 
     new javax.swing.JButton();
    button.addActionListener(this);
    return button; 
}
\end{codefootnote} \\ 
        \hline
    \end{tabular}

    \caption{Example synthesis outputs: (a) Writing to a file and (b) Creating a Swing JButton. For brevity, we omit the method body in the query, denoted as \textbf{...}. The method body was fed into transformers as a part of the prompt during generation.}
        \label{table:result_qual_supp1}
\end{table}

\begin{table}
    \centering
    \begin{tabular}{|p{1.35cm}|p{6cm}|p{6cm}|}
        \hline
    & (c) Connect to a Socket & (d) Decrypting a Message   \\ 
        \hline
    \scriptsize \textbf{ Query } &
        \vspace{-10 pt}
        \begin{codefootnote}
public class tcpSocketManager{

  public Socket createSocket(final 
     InetSocketAddress socketAddress){
     ...
  }

  public void connect(
     InetSocketAddress socketAddress){
  }
}
\end{codefootnote}
 
& \vspace{-10 pt} 
\begin{codefootnote}
public class lockAndKey{

 public String encryptMessage(
     String message, String secret_key){
        ...
 }

 public String decryptMessage(
     String encrypted_message, 
     String secret_key){
 }
}
\end{codefootnote}
 \\
        \hline
    \scriptsize \textbf{ \nag } &
        \vspace{-10 pt}
        \begin{codefootnote}
public void connect(
     InetSocketAddress fp_3)
  Socket var_8;
  var_8 = new Socket.Socket(
      (String) ARG,(int) ARG);
  try {
    InetSocketAddress var_3;
    var_3 = new InetSocketAddress.
       InetSocketAddress((String) ARG,
                         (int) ARG);
    var_8.connect((SocketAddress) var_3,
                  (int) ARG);
    }
  catch(ConnectException var_3){
        var_3.printStackTrace();
    }
  try {
        var_8.close();
    }
  catch(IOException var_3){
        var_3.printStackTrace();
    }
  return;
\end{codefootnote}
  
& \vspace{-10 pt}
    \begin{codefootnote}
String decryptMessage(String fp_2, 
                      String fp_5){
  if (fp_5.equals((Object) ARG)){
    return fp_5;
  }
  int var_6;
  var_6 = fp_5.indexOf((String) fp_5);
  if (true){
    return fp_5;
  }
  String var_4;
  var_4 = fp_5.substring((int) ARG,
                         (int) var_6);
  return var_4;
}
\end{codefootnote}
 \\
    \hline
    \scriptsize \textbf{ \codegpt } &
        \vspace{-10 pt}
        \begin{codefootnote}
public void connect(InetSocketAddress 
                    socketAddress)
  if (socketAddress ~= null) {
  socketAddress.connect(socketAddress);
}
\end{codefootnote} 
& \vspace{-10 pt}
  \begin{codefootnote}
public String decryptMessage(
           String encrypted_message,
           String secret_key){ 
return decryptMessage(encrypted_message,
                      secret_key);
}
\end{codefootnote}
 \\ 
        \hline
    \scriptsize \textbf{ \codex } &
        \vspace{-10 pt}
        \begin{codefootnote}
public void connect(InetSocketAddress 
                         socketAddress){
  AbstractSocket activeProxySocket = 
    createSocket(socketAddress);
  activeProxySocket.connect(
    socketAddress);    
}
\end{codefootnote} 
& \vspace{-10 pt}
  \begin{codefootnote}
public String decryptMessage(
           String encrypted_message,
           String secret_key){ 
 Key aesKey = new SecretKeySpec(
   secret_key.getBytes(), 'AES'); 
 Cipher cipher = 
   Cipher.getInstance('AES');
 cipher.init(Cipher.DECRYPT_MODE, aesKey); 
 return cipher.doFinal(encrypted_message);
}
\end{codefootnote} \\ 
        \hline
    \scriptsize \textbf{ GptNeo1.3B } &
        \vspace{-10 pt}
        \begin{codefootnote}
public void connect(InetSocketAddress 
                    socketAddress)
  throws IOException {
    Socket s = new Socket(
      new Proxy(
        Proxy.Type.SOCKS,socketAddress)
      );
  s.connect();
}
\end{codefootnote} 
& \vspace{-10 pt}
  \begin{codefootnote}
public String decryptMessage(
           String encrypted_message,
           String secret_key){ 
 Key aesKey=
   new SecretKeySpec(
     encrypted_message.getBytes(), 
     secret_key.getBytes()
    );
   return aesKey.doFinal(
     encrypted_message);
}
\end{codefootnote}
 \\ 
        \hline
    \end{tabular}

    \caption{Example synthesis outputs: (a) Connect to a socket and (b) Decrypting a message. For brevity, we omit the method body in the query, denoted as \textbf{...}. The method body was fed into transformers as a part of the prompt during generation.}
        \label{table:result_qual_supp2}
\end{table}

\section{Static Checks Considered} \label{appendix:semantic_properties}

Here we give an in-depth description of each of the static checks that has been tested in the paper as described in Section \ref{sec:quant_results}.

\begin{itemize}
    \item \textbf{No Undeclared Variable Access.} All the variables used in a program should be declared before they are used and should be available in the scope.
    We measure the percentage of variable usages across all the programs that are declared before use.
    For example \texttt{int bar() \{x.write();\}} is a violation of this property because \texttt{x} is not declared (assuming that there is no field named \texttt{x}).
    When the statement \texttt{x.write()} is synthesized, the \nag-model has access to the symbol table at that point in the AST.
    The symbol table does not contain the variable \texttt{x} because it is not declared in the preceding statements.
    The \nag-model has learnt to use variables that are present in the symbol table (encoded as attribute "\att{symTab}"), which increases the likelihood that the variables used in the program are declared before being used.
    \vspace{0.2cm}
    \item \textbf{Valid Formal Parameter Access.} All of the input variables accessed in a program body should be available in the class definition. Across all programs, we measure the percentage of input-variable accesses that are available. If a program grammar allows access for $n$ input variables, and the synthesizer tries to access one such variable $n_k$ even when it is not available, this property will be violated. The formal-parameter-type information is present in \emph{symTab} corresponding to each of the input variable which helps \nag learn this property correctly.
    \vspace{0.2cm}
    \item \textbf{Valid Class Variable Access.} All the class fields accessed in a program body should be available in  the class definition. The presence of field information in \emph{symtab} helps \nag satisfy this semantic property. Across all programs, we measure the percentage of field accesses that happened when they were available.
    \vspace{0.2cm}
    \item \textbf{No Uninitialized Objects.}
    All variables with reference type should be initialized.
    Out of all the variables with reference types declared across all the programs, we measure the percentage of variables that are initialized using a "\texttt{new}" statement. 
    For example, \texttt{BufferedWriter x;} \texttt{x.write();} is a violation because \texttt{x} is not initialized using \texttt{new BufferedWriter}.
    Violation of this property could cause a \emph{NullPointerException} at runtime.
    The AG keeps track of variable initializations using an attribute of type array of Booleans, named \emph{IsInitialized}.
    Whenever a variable is declared, the corresponding value in the \emph{IsInitialized} array is set to \emph{False}.
    As soon as the variable is initialized, the attribute is set to \emph{True}. This attribute helps the \nag model learn to avoid generating method bodies in which a variable is used without being initialized.
    \vspace{0.2cm}
    \item \textbf{No Variable Access Errors.} This property is the aggregate of the preceding four semantic checks. 
    \vspace{0.2cm}
    \item \textbf{Object-method compatibility.} Methods should be invoked on objects of appropriate types. For example,  consider the program snippet 
    \texttt{ int k = 0; \\bool b = k.startsWith(pre);} 
    which contains an invocation of  
    \texttt{String::startsWith(String pre)}. 
    This program fails the object-method-compatibility check because the method is invoked on a variable of type \emph{int} instead of a variable of type \emph{String}.
    The method invocation \emph{startsWith} is synthesized before the terms \emph{k} and \emph{pre}.\footnote{\label{footnote:MethodInvocationSynthesisOrder}Note that while this attribute grammar requires $\symb{Method}$ to be expanded before $\symb{Expr}$ (because inherited attributes of the latter depend on synthesized attributes of the former), the grammar is still $L$-attributed if we expand $\symb{Method}$ then $\symb{Expr}$, and perform an unparsing trick to emit the subtree produced by $\symb{Expr}$ first.}
    The symbol-table attribute of the AG, in combination with the synthesized attribute \emph{expr\_type}, helps the \nag model avoid such cases by learning to synthesize an expression with a compatible type, given the method signature to be invoked on it.
    %
    %
    \vspace{0.2cm}
    \item\textbf{Return Type at the Call Site.} The return type of a method invoked at some call site should be consistent with the type expected at the call site. In \texttt{ bool b = aStr.startsWith(pre);} this property asserts that the type of \texttt{b} should be compatible with the return type of \texttt{String::startsWith}. The symbol table alongside the synthesized attribute from the API call, namely \emph{ret\_type}, helps the \nag respect this rule.
    \vspace{0.2cm}
    \item \textbf{Actual Parameter Type.} The actual-parameter types in an API call should be consistent with the corresponding formal-parameter types.
    For example, consider the program fragrment
    \texttt{ int pre = 0; bool b = aStr.startsWith(pre);} 
    which contains an invocation of \texttt{String::startsWith(String pre)}.
    This program fails the formal-parameter type check because the actual-parameter type (\emph{int}) does not match the formal-parameter type (\emph{String}).
    The AG has the symbol-table attribute, which contains the variables in scope and their types, plus it has access to the intended type from the API call by the attribute \emph{typeList}, which helps the \nag model to learn to synthesize API arguments of appropriate types.
    \vspace{0.2cm}
    \item \textbf{Return Statement Type.} The type of the expression in a return statement of a method should be consistent with the method's declared return type.
    For example, \texttt{public int foo()\{String x; return x\}} violates this property because the returned expression \texttt{x} is of type \emph{String}, whereas the declared return type is \emph{int}.
    To make it easier for the \nag model to learn this restriction, the AG has a dedicated attribute \emph{methodRetType} to propagate the declared return type throughout the method, which helps it generate an expression of the appropriate type after consulting the symbol table. 
    For this property, we measure the percentage of return statements for which the expression type matches the method's declared return type.
    \vspace{0.2cm}
    \item \textbf{No Type Errors.} 
    All variables should be accessed in a type-consistent manner.
    Across all the programs, we measure the percentage of variable accesses that are type-consistent.
    Locations of variable accesses include all types of variable accesses relevant to an API call, method arguments, return statements, the variables on which the methods are invoked, variable assignments, and internal class method calls.
    \vspace{0.2cm}
    \item \textbf{Return Statement Exists.} 
    This property asserts that a method body should have a return statement. 
    \texttt{public int foo()\{String x;\}} violates this property because the method body does not have a return statement. 
    The AG propagates an attribute, \emph{retStmtGenerated}.
    This attribute is initially set to \emph{false}.
    When a return statement is encountered, the attribute is set to \emph{true}. 
    The \nag model learns to continue generating statements while this attribute is \emph{false}, and to stop generating statements in the current scope when the attribute is \emph{true}. 
    For this property, we report the percentage of programs synthesized with a \texttt{return} statement.
    \vspace{0.2cm}
    \item \textbf{No Unused Variables.} There should not be any unused variables (variables declared but not used) in the method body. 
    For example, in \texttt{public void reader()\{String x; String y; x=field1.read()\}}, the variable \texttt{y} is an unused variable.
    To keep track of the unused variables, we use a  boolean array attribute  \emph{isUsed}. Entries in array corresponding to the used variables are \emph{true} whereas all other entries are \emph{false}.  Out of all the programs synthesized, we report the percentage of variables declared which have been used inside the method body.
    \vspace{0.2cm}
    \item \textbf{Percentage of Parsing.} A parser for the Java language should be able to parse the synthesized programs.
    We use an open-source Java parser, called javalang \cite{javalang}, and check for the number of programs that parse.
    This test does not include static-semantic checks;
    it only checks if a generated program has legal Java syntax.
    Note that \nag, CNG, and GNN2NAG models are rule-based generation and they are bound to parse by definition.
    The pre-trained language models, however, are not guaranteed to produce programs that exactly follow the grammar definition.
    Therefore we capture all such instances that throw parsing exceptions and report the resulting numbers.
    \vspace{0.2cm}
    \item \textbf{Pass All Checks.}
    This property is the aggregate of all of the preceding checks.
\end{itemize}

\section{Implementation Details} \label{sec:implementation}

We now give a few details about how the distributions required to instantiate an \nag are implemented in our Java prototype. 

\noindent \textbf{Evidence Encoder}:
The evidences that we support as input to user context include class-level information (e.g., class name, Java-type information of the instance variables, and methods in the same class that have already been implemented); along with the information from the method header.

Each of these evidence types is encoded in a way appropriate to its domain.
The method header has a separate encoding for each of its components: return type, formal parameters, and method name.
The natural-language description available as Javadoc is also included. In total, there are seven kinds of evidence that we consider in our context. 

The evidences are encoded together as follows:
class and method names are split using camel case and delimiters;
the resulting elements are treated as natural-language keywords, and encoded as a set, using a single-layer feed-forward network.
The other evidences that are represented as sets and encoded by similar neural network. 
The type information of the class variables, formal parameters, and Javadoc are encoded as sequential data using a single-layered LSTM.
The surrounding method is encoded as a concatenation of the three components of the method header, namely, the method name, formal parameters, and return type, followed by a dense layer to reduce the dimensionality to the size of the latent space. 
Note that the model defined in Section \ref{sec:condproggen} allows us to get meaningful synthesis outputs even when only a small subset of the different kinds of evidence are available during training. 


\noindent \textbf{Sampling Symbol RHS Values}:
The distribution $P(\S | \mathsf{SymSoFar}, A(S){\downarrow}, \sZ)$
is implemented using an LSTM. 
There are six different kind of symbols for which we need to chooses an RHS: choosing the program block to produce (e.g., producing a try-catch statement or a loop), Java types, object-initialization calls, API calls, variable accesses, and accessing a method within the same class.
Each one of these has their own vocabulary of possibilities, and requires a separate neural LSTM decoding unit.
It is also possible to use additional, separate neural units in different production scenarios.
In our implementation, we use four separate LSTM units for decoding variable accesses: for a variable that is being declared, when accessed in a return statement, when accessed as an input parameter, or when an API call is invoked.
In other words, the \nag synthesizer consists of multiple decoding neural units, for decoding all of the production rules in the program's grammar, each using a separate LSTM unit.
It should be noted here that even though each of these LSTM units in the network has its own parameter set, they all maintain the same recurrent state, which tracks the state of the unfinished program synthesized so far. 


\noindent \textbf{Attributes}: Each of the neural units in an \nag decodes the current symbol using its corresponding LSTM and additional attributes available from the attribute grammar. Generally when a recurrent model like an LSTM is trained, the input to the LSTM cell is fixed as the correct output from the last time step (or the output from the parent node in case of a tree decoder). The availability of attributes in an \nag lets us augment this input information with the additional attributes from our grammar. The attributes that we support are given below:

\begin{itemize}
    \item Symbol table: An unrolled floating-point matrix that represents the types of all variables in scope, including field variables, input variables, and user-defined variables. Represented in our grammar as \textit{symTab} attrbute.
    \item Method return type: A floating-point vector containing the expected type of the method body.  Represented in our grammar as \textit{methodReturnType}.
    \item Return type of an API call, expression type of an object invoking an API call, and types of the input variables of an API call: Three separate floating-point vectors that represent the expected return type (\textit{retType}), the expression type of the object that initiates an API call (\textit{exprType}), and the expected formal parameters of the API call, if any (\textit{typeList}).
    
    
    \item Internal-method table: A floating-point vector representing the neural representation of the completed methods available in the same class.
    \item Unused variable flag: A Boolean vector indicating which variables have been initialized but not used so far in the program. The attribute that tracks this semantic property is \textit{isUsed}.
    \item Uninitiated-object flag: A Boolean vector indicating which objects have been declared but not initialized. The attribute that tracks this semantic property is \textit{isInitialized}.
    \item Return-statement flag: A Boolean indicating if a return statement has yet been reached in the program body. The attribute that tracks this semantic property is  \textit{retStmtGenerated}
\end{itemize}

Note that not all attributes are important to every production rule in the grammar at a given time step. For example, while decoding a variable access, it is unimportant to know about internal methods.
This information follows from our attribute grammar, as described in Appendix \ref{sec:grammar}.
If a particular attribute is not associated with a non-terminal, or it is unused inside a production rule, it is not required required for that rule, and the attribute can be omitted from being input to decoding that particular token.

\noindent \textbf{Training and Inference}: 
During training, all information required for the LSTM, such as contextual information for the missing class and the associated attributes in the AST, are available to the neural decoder.
The neural network is only tasked with learning the correct distributions related to decoding the method-body AST. 
The objective function related to learning the probability distributions within the learner is composed of a linear sum of cross-entropy loss for each category of symbol: non-terminals of the AST, API calls, Java types, and so on. This loss can be minimized using standard techniques like gradient descent to `train' the model.
If we compare this to a simple neural model, the \nag decoder has additional inputs in the form of attributes coming from the grammar to aid its decoding process.

 
During inference, only the program context is available to the decoder.  The attributes that the synthesizer requires are inferred on-the-fly, after the previous sub-part of the AST has been decoded.
Because we make use of an L-attributed grammar, at each step of AST construction, the necessary inputs are already available from the partial AST at hand.
At no point in the decoding process do the neural units need any input from the part of the AST that has not yet been synthesized.
This approach to synthesizing is close to the standard inference procedure in sequence-to-sequence models \cite{sutskever2014sequence, bert} 

Given the learned neural units, decoding for the ``best'' program is an intractable problem, hence we use beam search  \cite{vaswani2017attention}.
Because beam search is only applicable for sequences, we modify it to perform a depth-first traversal on the AST, with the non-terminal nodes that lead to branching in the AST stored in a separate stack.

\section{Implementation of Baselines}
\label{appendix:transformer}


\textbf{GNN2NAG}: We describe the implementation details of GNN2NAG~\citep{brockschmidt2018generative}, which is one of the baselines in Sec.~\ref{sec:eval}. We expand the AST nodes in the order of a depth-first search.
We considered the same six types of edges as~\citet{brockschmidt2018generative}, which consists of \emph{Parents}, \emph{Child}, \emph{NextSib}, \emph{NextUse}, \emph{NextToken}, and \emph{InhToSyn}.
After building the graph, we propagate the information through a Gated Graph Neural Network (GGNN~\citep{Li2016GatedGS,brockschmidt2018generative}). We obtain a representation for each node after the GGNN propagation.
We then apply an LSTM to go over all the nodes until reaching the node where the goal is to predict the next token. Training is via cross-entropy loss. Note that the biggest difference between our implementation of GNN2NAG and~\citet{brockschmidt2018generative} is the use of an LSTM.
~\citet{brockschmidt2018generative} assumes that information about which type of edge is responsible for generating the token is available to the model. However, this information is not available in our setup. Thus, we use an LSTM to iterate over all edges in the GNN to obtain the features for prediction.

\textbf{Pre-Trained Language Models}: 
We consider 4 types of transformer models---GPTNeo 125M, GPTNeo 1.3B, \codegpt, and \codex~\citep{gpt-neo, lu2021codexglue, codex}. We fine-tuned each of these pre-trained transformers on our Java dataset, except for \codex, for which we have no access to the pre-trained weights.
While our \nag model only takes the headers in the Java class as inputs, for the various transformer models, the input is the entire Java class, including both headers and method bodies.
(We found that transformers perform quite poorly if only headers are provided.)
During evaluation, transformers are asked to predict the missing method body given the header and the rest of the class.

To fine-tune on our Java dataset, we used the token-level code-completion task provided by CodeXGLUE\footnote{\url{https://github.com/microsoft/CodeXGLUE/tree/main/Code-Code/CodeCompletion-token}}~\citep{lu2021codexglue}. During fine-tuning, the transformers are asked to predict the next token in an autoregressive fashion, just like in any language-modeling task. The learning rate is $8e{-5}$ and the batch size is 1 on each GPU. In total, we used 16 GPUs to fine-tune these transformers, which takes about 4 days to complete two epochs on our Java dataset, consisting of 600,000 training instances.

\section{Generation of Training Data}

Now we sketch the process by which our training data is generated. 
Assume that the task is to generate procedure bodies from start-nonterminal $\symb{Prog}$, and that we are given a large corpus of Java programs from which to learn the distribution $P(\symb{Prog} | \sX)$. 


An AG-based compiler is used to produce the training data.
For each user-defined method $M$ in the corpus, we create training examples of the form
\begin{equation}
  \label{eqn:training}
  \left((\symb{Prog}, \sS_1), ..., (\S_{i-1}, \sS_{i-1}), (\S_{i}, \sS_{i}), A(\S_i){\downarrow}, \sX \right)
\end{equation}
where (i) $(\symb{Prog}, \sS_1), ..., (\S_{i-1}, \sS_{i-1})$ is a pre-order listing---from goal nonterminal $\symb{Prog}$ to a particular instance of nonterminal $\S_{i}$---of the (nonterminal, RHS) choices in $M$'s $\symb{Prog}$ subtree,
(ii) $\sS_{i}$ is the RHS production that occurs at $\S_{i}$, and
(iii) attribute values $A(\S_i){\downarrow}$ are the values at the given instance of $\S_{i}$.
As input-output pairs for a learner, inputs (i) and (iii) produce output (ii).

We compile the program, create its parse tree, and label each node with the values of its attributes (which are evaluated during a left-to-right pass over the tree).
For each method $M$, its subtree is traversed, and a training example is emitted for each node of the subtree.

\section{Restricting Available Evidence} \label{sec:25_evidence_experimeents}
In our experiments, generation of a particular method is conditioned on available ``evidences,'' which refer to the context surrounding the missing method, in the method's complete class  (other method names and method headers, Java Doc comments, class variables, and so on).  
All of the experiments described thus far simulate the situation where the entire class---except the method to be generated---is visible when it is time to generate the missing method.  This simulates the situation where a user is using an automatic programming tool to help generate the very last method in a class, when all other methods and class variables have been defined and are visible.

We can restrict the amount of evidence available to make the task more difficult.  When we only make a portion of the evidence available, this simulates the case where a user is using an automatic programming tool to generate a method when the surrounding class is less complete.
When we use ``$x$\% evidence'' for a task, each piece of evidence in the  surrounding  code is selected and available to the automatic programming tool with $x$\% probability.  In Table \ref{table:result_semantic_check_category_25pc} and Table \ref{table:max_topk_jaccard_25pc},  we show results obtained when we repeat the experiments from earlier in the paper, but this time using  25\% evidence while Table  \ref{table:result_semantic_check_category_50pc} and Table \ref{table:max_topk_jaccard_50pc} show the result for 50\% evidence.



\begin{table*}[t] 
	\centering
	\vspace{-8 pt}
	\caption{ Percent of Static Checks Passed with 25\% Evidence}
	\small
	\resizebox{\columnwidth}{!}{%
	\begin{tabular}{|l|c|c|c|c|c|c|c|c|}
		\hline
		\textit & GPTNeo125M & GPTNeo1.3B & CodeX & \codegpt & GNN2NAG & CNG & \nag  \\
		\hline
		\hline
		No Undeclared Variable Access    & 90.77\% &  89.99\% & 84.55\% & 89.21\%  & 46.88\% & {19.78\%} & \textbf{99.32\%}\\
		\hline
		Valid Formal Param Access    & NA & NA & NA & NA &   25.72\% & {11.03\%} & \textbf{98.61\%}\\
		\hline
		Valid Class Var Access    &  NA & NA & NA & NA  & 14.34\% & {12.75\%} & \textbf{99.31\%} \\
		\hline
		No Uninitialized Objects    &  92.35\% & 91.21\% & 88.52\%  & \textbf{93.40\%}  & 20.31\% &{21.56\%} & 93.35\% \\
		\hline
        \hline
        No Variable Access Error    & 90.92\% & 90.11\% & 84.98\% & 89.63\%  & 28.10\% & 17.92\% & \textbf{99.10\%}  \\
        \hline
        \hline
        Object-Method Compatibility & {96.93\%} & 97.05\% & 96.74\% & \textbf{98.35\%}  & 21.29\% & 12.23\% & 94.87\%  \\
        \hline
        Ret Type at Call Site     & 97.91\% & 97.66\% & \textbf{98.47\%} & 97.98\%  & 22.97\% & 16.40\% & 92.53\% \\
        \hline
        Actual Param Type       & 88.51\% & 88.61\% & 90.02\% & 86.77\%  & 9.22\% & 16.09\% & \textbf{93.63\%} \\
        \hline
        Return Stmt Type         & 82.50\% & 81.56\% & 83.75\% & 83.43\%  & 12.05\% & 9.51\%  & \textbf{88.94\%} \\
        \hline
        \hline
        No Type Errors      & 86.46\% & 86.14\% & 88.62\% & 86.83\% & 15.98\% & 13.56\%  & \textbf{91.98\%} \\
        \hline
        \hline
        Return Stmt Exists  & 99.58\% & 99.61\% & 96.74\% & 99.56\%  & 93.83\% & \textbf{99.92\%}  & {96.94\%} \\
        \hline
        No Unused Variables     & 96.51\% & 96.33\% & 96.46\% & \textbf{97.60\%}  & 20.14\% & 24.29\% & 91.75\%  \\
        \hline
        \hline
         Percentage of Parsing   & 98.61\% & 98.53\% & 94.95\% & 97.14\% & \textbf{100.0}\% & \textbf{100.0}\% & \textbf{100.0}\% \\
         \hline
         Pass All Checks  &  65.26\% & 62.65\% & 38.77\% & 63.12\% & 16.75\% & 12.87\% & \textbf{79.17\%} \\
         \hline
	\end{tabular}
	}
	\label{table:result_semantic_check_category_25pc}
\end{table*}

\begin{table*}[t]
	\centering
	\caption{ Average Fidelity of Generated Method Bodies with 25\% Evidence}
	\small
	\resizebox{\columnwidth}{!}{%
	\begin{tabular}{|l|c|c|c|c|c|c|c|}
	    \hline
		\textit & GPTNeo125M & GPTNeo1.3B & CodeX & \codegpt  & CNG  & \nag \\
		\hline
		\hline
		Set of API Calls     &  24\% & 27\% & 29\%  & 27\% & 12\% & \textbf{43\%} \\
		\hline
		Sequences of API Calls  &  12\% & 14\% & 13\% & 13\% & 7\%  &  \textbf{31\%}  \\
		\hline
        Sequences of Program Paths     & 7\% & 8\% & 8\% & 8\% & 7\% &\textbf{28\%} \\
        \hline
        AST Exact Match                &  7\% & 8\% & 8\% & 8\% & 1\% & \textbf{18\%} \\
        \hline
	\end{tabular}
    }
	\label{table:max_topk_jaccard_25pc}
\end{table*}

\begin{table*}[t] 
	\centering
	\vspace{-8 pt}
	\caption{ Percent of Static Checks Passed with 50\% Evidence} 
	\small
	\resizebox{\columnwidth}{!}{%
	\begin{tabular}{|l|c|c|c|c|c|c|c|c|}
		\hline
		\textit & GPTNeo125M & GPTNeo1.3B & CodeX & \codegpt & GNN2NAG & CNG & \nag  \\
		\hline
		\hline
		No Undeclared Variable Access    & 89.87\% &  90.36\% & 88.62\% & 90.34\%  & 47.17\% & {17.79\%} & \textbf{99.86\%}\\
		\hline
		Valid Formal Param Access    & NA & NA & NA & NA &   25.50\% & {8.58\%} & \textbf{99.83\%}\\
		\hline
		Valid Class Var Access    &  NA & NA & NA & NA  & 14.96\% & {11.57\%} & \textbf{99.78\%} \\
		\hline
		No Uninitialized Objects    &  93.90\% & 91.73\% & 90.82\%  & 94.37\%  & 20.01\% &{21.68\%} & \textbf{97.30\%} \\
		\hline
        \hline
        No Variable Access Error    & 90.36\% & 90.51\% & 88.86\% & 91.32\%  & 28.43\% & 17.34\% & \textbf{99.84\%}  \\
        \hline
        \hline
        Object-Method Compatibility & \textbf{98.36\%} & 98.09\% & 98.35\% & 97.84\%  & 21.39\% & 10.11\% & 96.42\%  \\
        \hline
        Ret Type at Call Site     & 97.38\% & 98.01\% & \textbf{98.53\%} & 97.83\%  & 23.45\% & 14.82\% & 97.22\% \\
        \hline
        Actual Param Type       & 87.03\% & 86.36\% & 92.28\% & 88.71\%  & 9.24\% & 14.35\% & \textbf{96.74\%} \\
        \hline
        Return Stmt Type         & 84.05\% & 85.09\% & 88.13\% & 85.23\%  & 12.07\% & 7.66\%  & \textbf{92.15\%} \\
        \hline
        \hline
        No Type Errors      & 87.25\% & 88.13\% & 91.42\% & 88.10\% & 16.04\% & 11.45\%  & \textbf{96.22\%} \\
        \hline
        \hline
        Return Stmt Exists  & 99.61\% & 99.80\% & 98.44\% & \textbf{99.57\%}  & 93.87\% & {98.71\%}  & {97.47\%} \\
        \hline
        No Unused Variables     & 96.42\% & 96.46\% & 96.82\% & \textbf{97.64\%}  & 20.55\% & 18.50\% & 94.20\%  \\
        \hline
        \hline
         Percentage of Parsing   & 98.18\% & 98.13\% & 94.69\% & 97.08\% & \textbf{100.0}\% & \textbf{100.0}\% & \textbf{100.0}\% \\
         \hline
         Pass All Checks  &  65.26\% & 64.88\% & 47.49\% & 67.73\% & 16.92\% & 24.28\% & \textbf{86.00\%} \\
         \hline
	\end{tabular}
	}
	\label{table:result_semantic_check_category_50pc}
\end{table*}

\begin{table*}[t]
	\centering
	\caption{ Average Fidelity of Generated Method Bodies with 50\% Evidence}
	\small
	\resizebox{\columnwidth}{!}{%
	\begin{tabular}{|l|c|c|c|c|c|c|c|}
	    \hline
		\textit & GPTNeo125M & GPTNeo1.3B & CodeX & \codegpt  & CNG  & \nag \\
		\hline
		\hline
		Set of API Calls     &  32\% & 37\% & 36\%  & 36\% & 12\% & \textbf{50\%} \\
		\hline
		Sequences of API Calls  &  17\% & 20\% & 16\% & 19\% & 7\%  &  \textbf{39\%}  \\
		\hline
        Sequences of Program Paths    & 13\% & 10\% & 10\% & 14\% & 7\% &\textbf{36\%} \\
        \hline
        AST Exact Match                &  13\% & 10\% & 10\% & 14\% & 1\% & \textbf{21\%} \\
        \hline
	\end{tabular}
    }
	\label{table:max_topk_jaccard_50pc}
\end{table*}

\section{Next-Token Prediction} \label{sec:appendix_nexttoken}

Our \nag implementation uses a relatively weak language model (based on LSTMs as opposed to more modern transformers) but augments them with a static analysis.  We have shown that the resulting \nag is good at ``long-horizon'' tasks such as semantic consistency  (compared to the baselines tested) and at generating methods that have high fidelity to the original, ``correct'' method.  But it is reasonable to ask: how does the \nag compare to the baselines at ``short-horizon'' tasks?
To measure this, for each symbol $S$ that is expanded to form the body of a test method, we compute (i) the actual left-context sequence of the test method (up to but not including the RHS sequence chosen for $S$) as the value of SymSoFar, (ii) $A(S) \downarrow $ (in the case of the \nag), and (iii) $\sZ$. We then use these values to ask the \nag to predict the next RHS. If the predicted RHS matched the observed RHS, the model was scored as “correct.” We recorded the percentage of correct predictions for terminal RHS symbols (such as API calls or types) for each test program.

We also performed next-token prediction using the \nag and three of the baseline models.
Note that it is non-trivial to classify \codegpt's output into different terminal symbols, so we only report the overall RHS symbols' correctness.
The results show that two of the baselines (\codegpt and GNN2NAG) are very accurate, and demonstrate better performance than the \nag on this task.
These results are in-keeping with our assertion that the baselines are useful mostly for short-horizon code-generation tasks.
However, they struggle with long-horizon tasks, such as the CPG task of generating an entire Java method body.
The results---together our earlier CPG results---also show that even though the \nag has reduced accuracy in a short-horizon task, it is still able to generate semantically accurate programs on the CPG task.

\begin{table*}[t]
	\centering
\vspace{-13 pt}
	\caption{ Next-Token Prediction Accuracy }
	\small
	\resizebox{\columnwidth}{!}{
	\begin{tabular}{|l|c|c|c|c|c|c|c|c|}
	    \hline
	    & 	\multicolumn{8}{c|}{\textbf{Percentage of Evidence Available}}\\
		\hline
		& \multicolumn{4}{c|}{\textbf{50\%}}  & \multicolumn{4}{c|}{\textbf{100\%}} \\
		\hline
		\textit & \nag & \codegpt & GNN2NAG & CNG  & \nag & \codegpt  & GNN2NAG & CNG   \\
		\hline
		\hline
		API Calls                       & 62.42\% & NA & \textbf{80.24\%} & 49.05\% &  75.94\% & NA & \textbf{80.77\%} & 59.73\% \\
		\hline
		Object Initialization Call  & 59.64\% & NA & \textbf{97.65\%} & 49.12\% & 66.66\% & NA & \textbf{97.94\%} & 87.90\% \\
        \hline
        Types                           & \textbf{61.11\%} & NA & \textbf{85.78\%} & 50.28\% &  70.33\% & NA & \textbf{86.21\%} & 54.44\%    \\
        \hline
        Variable Access                 & \textbf{92.26\%} & NA & 92.11\%  & 50.28\% &  92.44\% & NA & \textbf{92.94\%} & 52.85\%  \\
        \hline

        \hline
        All Terminal RHS Symbols    & 73.41\% & \textbf{88\%} & 80.83\%  & 51.22\% & 73.99\% & \textbf{89\%} & 81.1\% & 54.32\%  \\
        \hline
	\end{tabular}
	}
	\label{table:result_relevance_check}
\end{table*}
\section{Application to Novel Semantic Checks} \label{sec:ablation_study}

The \nag approach can generate semantically accurate programs given  context.
At its core, an \nag relies on the various semantic properties (i.e., attributes) on which it is trained. We would like to understand the influence of these semantic properties in the generated program, and explore the possibility that training on such a set of attributes can automatically allow for high accuracy with respect to additional semantic checks for which specific attributes were not explicitly provided during training.
To study this question, we performed an ablation study in which we trained an \nag with a subset of the relevant attributes, but evaluated the generated programs on all properties. 

We trained an \nag without the $\att{\textit{attrOut.retStmtGenerated}}$ and $\att{\textit{methodRetType}}$ attributes, as defined in  Section \ref{appendix_grammar:attribute_grammar}. With 50\% of the evidence available, we see that the resulting model suffers in terms of accuracy. The ``Return Stmt Type'' accuracy falls from 92.15\% to 77.45\% whereas the ``Return Stmt Exists" accuracy falls from 97.47\% to 95.68\%. That said, note that the resulting ``Return Stmt Type'' accuracy is still a big improvement over the vanilla CNG model (with no attributes), which is correct only 9.51\% of the time. 

This suggests that the \nag has learned type-safe program generation from other semantic properties, most notably the $\att{\textit{symTab}}$ attribute, which carries type information about the various objects that are currently in scope.  This further suggests that providing a small core of key attributes may be enough to greatly increase the accuracy of code generation.





\section{Robustness Incomplete Analysis} \label{sec:appendix_broken_compiler}


In this section, we analyze a situation where the static analyzer fails to accurately resolve different attributes during the synthesis process. We simulate three situations in which the static analyzer might fail.

In the first scenario, we emulate a situation where the compiler is unable to resolve the correct return-type information from the missing method that the user has asked the \nag to synthesize. This results in a default \texttt{null} value being passed around for the attribute $\att{\textit{methodRetType}}$. We find that this reduces the overall accuracy for the attribute ``Return Stmt Type"  from 90.97\% to 77.28\%. This does not seem to impact other static checks, however.



In the second scenario, consider the case where the compiler is unable to resolve the API return-type attribute $\att{\textit{retType}}$.
This reduces the accuracy of the ``Return type at call site'' check from 98.01\% to 18.16\%.
It also results in a decrease in ``No undeclared-variable access'' and ``Valid formal-param access'' to 72.48\% and 67.23\%, respectively. This is is a huge decrease from 99.82\% and 99.55\% accuracy that these semantic checks had for the base model where the attribute $\att{\textit{retType}}$ can be resolved correctly.
The fidelity metrics are also impacted, where the ``AST exact match'' metric drops from 26\% to 10\%.
This is because the $\att{\textit{retType}}$ attribute is used in many portions of our attribute grammar, on which the trained program generator is being conditioned on. An incorrect resolution of such attribute had led to deterioration of the overall model performance.

In the final scenario, we only break the static analyzer's capability to resolve the unused-variable-check attribute $\att{\textit{attrOut.isUsed}}$. For this scenario, we see that only the one semantic check ``No unused variables" out of all the semantic checks considered is impacted. Here the accuracy for this check drops from 93.84\% to 91.10\%. All other metrics have negligible changes. 

Rather unsurprisingly, the results suggest that \nag relies heavily on the static analyzer, and that some attributes influence the result much more than others. It is also critical to have a static analyzer that performs accurately during inference time, to avoid any model performance degradation.


\section{BLEU-Score Analysis}
\label{appendix:bleu}

As described in the main body of the paper, BiLingual Evaluation Understudy or BLEU score is ``problematic in the code-generation setting.
First, the BLEU score is not invariant to variable renamings, which means that a nonsensical program that uses commonplace variable names can get an artificially high BLEU score. Second, programs are structured objects in which some tokens indicate control flow, and some indicate data flow. The BLEU score does not take this structure into account.'' We use one of the examples in Table \ref{table:result_qual_reader} of the paper ("reading from a file") to illustrate this point and show the real code and outputs from CodeGPT and \nag in Table \ref{table:result_bleu}. 

The \nag output is clearly better. However, the CodeGPT output gets a higher BLEU score because it uses variable names that superficially match the ground truth. Specifically, the BLEU score of the CodeGPT output is 25.11 and the BLEU score of the \nag output is 19.07. This situation arose often in our experiments, which is why we have used alternative program-equivalence metrics to judge performance of generated programs, as defined in Section \ref{sec:eval}.

\begin{table}
    \centering
    \begin{tabular}{|p{4.7cm}|p{4.0cm}|p{4.0cm}|}
        \hline
        \small Real Code & \small CodeGPT & \small \nag \\
        \hline
        \vspace{-10 pt}
\begin{codefootnote}
public String reader()
{
  StringBuffer stringBuffer 
            = new StringBuffer(); 
  String line; 
  while ((line =
  bReader.readLine() ~= null) {
    stringBuffer.append(line);
    stringBuffer.append("\n");};
  return stringBuffer.toString();
}
\end{codefootnote}
        & \vspace{-10 pt} 
\begin{codefootnote}
public String reader()
{
 StringBuffer buffer=
       new StringBuffer();
 buffer.append("\n");
 return buffer.toString();
}
\end{codefootnote}
& \vspace{-10 pt} 
        \begin{codefootnote}
public String reader()
{
  java.lang.String var_9; 
  try{
   var_9=field_5.readLine();
   } 
  catch(IOException var_8) {
    var_8.printStackTrace();
  }
  return var_9;
}
\end{codefootnote}
        \\
        \hline
    \end{tabular}
        \vspace{-5 pt}
    \caption{Reader example for analyzing the BLEU-score metric.}
  \vspace{-5 pt}
        \label{table:result_bleu}
\end{table}






\section{Grammar} \label{sec:grammar}

The Neural Attribute Grammar (\nag) model learns to synthesize real-life Java programs while learning over production rules of an attribute grammar. In this section, we present the comprehensive set of production rules considered, along with the attributes used.
We first present the context-free grammar in Appendix \ref{appendix_grammar:cfg}, and then decorate it with attributes in Appendix \ref{appendix_grammar:attribute_grammar}.
The productions in \textbf{a-c} deal with expansion of all the non-terminal symbols in the grammar:
rules in \textbf{a} mainly expand to one line of code in a Java method body;
rules in \textbf{b} are their corresponding expansions; and
rules in \textbf{c} deal with control-flow operations inside the grammar.
Rules in \textbf{d} generate terminal symbols inside the grammar.
We show the flow of attributes \textit{symTab} and \textit{methodRetType} in the AST in Appendix \ref{appendix_grammar:attribute_grammar}. The rest of the attributes are passed inside \textit{attrIn} and \textit{attrOut}, namely \textit{isInitialized}, \textit{isUsed}, \textit{retStmtGenerated} and \textit{itrVec}. 

\subsection{Context Free Grammar} \label{appendix_grammar:cfg}
\begingroup 
\small
$\begin{array}{lll}
\textbf{a1.} & \symb{Start} : & \symb{Stmt} \\
\textbf{a2.} & \symb{Stmt} : & \symb{Stmt}~;~\symb{Stmt}~|~\epsilon \\
\textbf{a3.} & \symb{Stmt} : & \symb{Decl} \\
\textbf{a4.} & \symb{Stmt} : & \symb{ObjInit} \\
\textbf{a5.} & \symb{Stmt} : & \symb{Invoke} \\
\textbf{a6.} & \symb{Stmt} : & \symb{Return} \\

\end{array}$
\par\noindent
$\begin{array}{lll}
\textbf{b1.} & \symb{Decl} :       & \symb{Type}~\symb{Var}~ \\
\textbf{b2.} & \symb{ObjInit} :    & \symb{Type}~\symb{Var}~\textbf{=}~\textbf{new}~\symb{Type}~\symb{ArgList} \\
\textbf{b3.} & \symb{Invoke} :     & \symb{Var}~=~\symb{Var}~\symb{Call}~\symb{InvokeMore}~ \\
\textbf{b4.} & \symb{InvokeMore} : & \symb{Call}~\symb{InvokeMore}~|~\epsilon \\
\textbf{b5.} & \symb{Call} :       & \symb{Api}~\symb{ArgList} \\
\textbf{b6.} & \symb{ArgList} :    & \symb{Var}~\symb{ArgList}~|~\epsilon \\
\textbf{b7.} & \symb{Return} :     & \textbf{return}~\symb{Var} \\
\end{array}$

\par\noindent

$\begin{array}{lll}
\textbf{c1.} & \symb{Stmt} : & \symb{Branch}~|~\symb{Loop}~|~\symb{Except} \\
\textbf{c2.} & \symb{Branch} : & \ifc~\symb{Cond}~\thenc~\symb{Stmt}~\elsec~\symb{Stmt} \\
\textbf{c3.} & \symb{Loop} : & \whilec~\symb{Cond}~\thenc~\symb{Stmt} \\
\textbf{c4.} & \symb{Except} : & \textbf{try}~\symb{Stmt}~\symb{Catch} \\
\textbf{c5.} & \symb{Catch} : & \textbf{catch}(\symb{Type})~\symb{Stmt};~\symb{Catch}~|~\epsilon \\
\textbf{c6.} & \symb{Cond} : & \symb{Call} \\
\end{array}$
\endgroup 

\par\noindent
$\begin{array}{lll}
\textbf{d1.} & \symb{Api} : & \textbf{JAVA\_API\_CALL} \\
\textbf{d2.} & \symb{Api} : & \textbf{INTERNAL\_METHOD\_CALL} \\
\textbf{d3.} & \symb{Type} : & \textbf{JAVA\_TYPE} \\
\textbf{d4.} & \symb{Var} : & \textbf{VAR\_ID} \\
\end{array}$

\subsection{Attribute Grammar} \label{appendix_grammar:attribute_grammar}
\begingroup 
\small

$\begin{array}{ll}
\textbf{a0.} & \text{Initialization of inherited attributes of \symb{Start}:} \\
& \big[ ~\symb{Start}.\att{symTab} \downarrow ~:=~ \\
& \quad\{~ in\_param\_1 \mapsto type\_in\_param\_1, \\
& \quad ~~ \ldots \\
& \quad~~ in\_param\_n \mapsto type\_in\_param\_n, \\
& \quad~~ field\_1 \mapsto type\_field\_1, \\
& \quad ~~ \ldots \\
& \quad~~ field\_m \mapsto type\_field\_m ~\} \\
& \symb{Start}.\att{attrIn.itrVec} \downarrow ~:=~ \text{(false, false)}; \\
& \symb{Start}.\att{attrIn.retStmtGenerated} \downarrow ~:=~ \text{false}; \\
& \symb{Start}.\att{attrIn.isInitialized} \downarrow ~:=~ \phi; \\
& \symb{Start}.\att{attrIn.isUsed} \downarrow ~:=~ \phi; \\
& \symb{Start}.\att{methodRetType} \downarrow ~:=~ METHOD\_RET\_TYPE ; ]\\
\end{array}$
\par\noindent
$\begin{array}{ll}
\textbf{a1.} & \symb{Start}  : \symb{Stmt} ~; \\
& \big[~\symb{Stmt}.\att{attrIn} \downarrow ~:=~ \symb{Start}.\att{attrIn} \downarrow; \\
& \symb{Stmt}.\att{methodRetType} \downarrow ~:=~ \symb{Start}.\att{methodRetType} \downarrow; \\
& \symb{Stmt}.\att{symTab} \downarrow ~:=~ \symb{Start}.\att{symTab} \downarrow ;\\
& \symb{Start}.\att{symTabOut} \uparrow ~:=~ \symb{Stmt}.\att{symTabOut} \uparrow ; \\
& \symb{Start}.\att{attrOut} \uparrow ~:=~ \symb{Stmt}.\att{attrOut} \uparrow; \\
& \symb{Start}.\att{valid} \uparrow ~:=~ \symb{Stmt}.\att{valid} \uparrow; \big] \\
\end{array}$

\par\noindent
$\begin{array}{ll}
\textbf{a2a.} & \symb{Stmt}\$0  :  \symb{Stmt}\$1~;~\symb{Stmt}\$2 \\
& \big[~\symb{Stmt}\$1.\att{symTab} \downarrow ~:=~ \symb{Stmt}\$0.\att{symTab} \downarrow ;\\
& \symb{Stmt}\$2.\att{symTab} \downarrow ~:=~ \symb{Stmt}\$1.\att{symTabOut} \uparrow ;\\
& \symb{Stmt}\$0.\att{symTabOut} \uparrow ~:=~ \symb{Stmt}\$2.\att{symTabOut} \uparrow ;\\
& \symb{Stmt}\$1.\att{attrIn} \downarrow ~:=~ \symb{Stmt}\$0.\att{attrIn} \downarrow; \\
& \symb{Stmt}\$2.\att{attrIn} \downarrow ~:=~ \symb{Stmt}\$1.\att{attrOut} \uparrow; \\
& \symb{Stmt}\$0.\att{attrOut} \uparrow ~:=~ \symb{Stmt}\$2.\att{attrOut} \uparrow;  \\
& \symb{Stmt}\$1.\att{methodRetType} \downarrow ~:=~ \symb{Stmt}\$0.\att{methodRetType} \downarrow ;\\
& \symb{Stmt}\$2.\att{methodRetType} \downarrow ~:=~ \symb{Stmt}\$0.\att{methodRetType} \downarrow ;\\
& \symb{Stmt}\$0.\att{valid} \uparrow ~:=~ \symb{Stmt}\$1.\att{valid} \uparrow  \wedge~ \symb{Stmt}\$2.\att{valid} \uparrow; \big] \\
\end{array}$
\par\noindent

$\begin{array}{ll}
\textbf{a2b.} & \symb{Stmt}  :  \epsilon \\
& \big[~\symb{Stmt}.\att{symTabOut} \uparrow ~:=~ \{ \}; \\
& \symb{Stmt}.\att{attrOut.itrVec} \uparrow ~:=~ \text{(false, false)}; \\
& \symb{Stmt}.\att{valid} \uparrow ~:=~ \text{true}; \big] \\
\end{array}$

\par\noindent
$\begin{array}{ll}
\textbf{a3.} & \symb{Stmt}  :  \symb{Decl}  \\
& \big[~\symb{Decl}.\att{symTab} \downarrow ~:=~ \symb{Stmt}.\att{symTab} \downarrow ;\\
& \symb{Stmt}.\att{symTabOut} \uparrow ~:=~ \symb{Stmt}.\att{symTab} \downarrow +~ \symb{Decl}.\att{symTabOut} \uparrow ; \\
& \symb{Decl}.\att{attrIn} \downarrow ~:=~ \symb{Stmt}.\att{attrIn} \downarrow; \\
& \symb{Stmt}.\att{attrOut} \uparrow ~:=~ \symb{Stmt}.\att{attrIn} \downarrow +~ \symb{Decl}.\att{attrOut} \uparrow; \\
& \symb{Decl}.\att{methodRetType} \downarrow ~:=~ \symb{Stmt}.\att{methodRetType} \downarrow ;\\
& \symb{Stmt}.\att{valid} \uparrow ~:=~ \symb{Decl}.\att{valid} \uparrow; \big] \\
\end{array}$
\par\noindent

$\begin{array}{ll}
\textbf{a4.} & \symb{Stmt} : \symb{ObjInit} \\
& \big[~\symb{ObjInit}.\att{symTab} \downarrow ~:=~ \symb{Stmt}.\att{symTab} \downarrow ;\\
& \symb{Stmt}.\att{symTabOut} \uparrow ~:=~ \symb{Stmt}.\att{symTab} \downarrow \\
& \qquad \qquad +~ \symb{ObjInit}.\att{symTabOut} \uparrow ; \\
& \symb{ObjInit}.\att{attrIn} \downarrow ~:=~ \symb{Stmt}.\att{attrIn} \downarrow; \\
& \symb{ObjInit}.\att{methodRetType} \downarrow ~:=~ \symb{Stmt}.\att{methodRetType} \downarrow ;\\
& \symb{Stmt}.\att{attrOut} \uparrow ~:=~ \symb{Stmt}.\att{attrIn} \downarrow +~ \symb{ObjInit}.\att{attrOut} \uparrow; \\
& \symb{Stmt}.\att{valid} \uparrow ~:=~ \symb{ObjInit}.\att{valid} \uparrow \big] \\
\end{array}$

\par\noindent
$\begin{array}{ll}
\textbf{a5.} & \symb{Stmt} : \symb{Invoke} \\
& \big[~\symb{Invoke}.\att{symTab} \downarrow ~:=~ \symb{Stmt}.\att{symTab} \downarrow ;\\
& \symb{Stmt}.\att{symTabOut} \uparrow ~:=~ \symb{Stmt}.\att{symTab} \downarrow + \symb{Invoke}.\att{symTabOut} \uparrow ; \\
& \symb{Invoke}.\att{attrIn} \downarrow ~:=~ \symb{Stmt}.\att{attrIn} \downarrow; \\
& \symb{Stmt}.\att{attrOut} \uparrow ~:=~ \symb{Stmt}.\att{attrOut} \downarrow + \symb{Invoke}.\att{attrOut} \uparrow;  \\
& \symb{Invoke}.\att{methodRetType} \downarrow ~:=~ \symb{Stmt}.\att{methodRetType} \downarrow ;\\
& \symb{Stmt}.\att{valid} \uparrow ~:=~ \symb{Invoke}.\att{valid} \uparrow; \big] \\
\end{array}$
\par\noindent

\par\noindent
$\begin{array}{ll}
\textbf{a6.} & \symb{Stmt} : \symb{Return} \\
& \big[~\symb{Return}.\att{symTab} \downarrow ~:=~ \symb{Stmt}.\att{symTab} \downarrow ;\\
& \symb{Stmt}.\att{symTabOut} \uparrow ~:=~ \\
& \qquad \symb{Stmt}.\att{symTab} \downarrow +~ \symb{Return}.\att{symTabOut} \uparrow ; \\
& \symb{Invoke}.\att{attrIn} \downarrow ~:=~ \symb{Stmt}.\att{attrIn} \downarrow; \\
& \symb{Stmt}.\att{attrOut} \uparrow ~:=~ \symb{Stmt}.\att{attrIn} \downarrow +~ \symb{Invoke}.\att{attrOut} \uparrow; ] \\
& \symb{Return}.\att{methodRetType} \downarrow ~:=~ \symb{Stmt}.\att{methodRetType} \downarrow ;\\
& \symb{Stmt}.\att{valid} \uparrow ~:=~ \symb{Return}.\att{valid} \uparrow; \big] \\
\end{array}$
\par\noindent

$\begin{array}{ll}
\textbf{b1.} & \symb{Decl} : \symb{Type}~\symb{Var}~ \\
& \big[\symb{Decl}.\att{symTabOut} \uparrow ~:=~ \{ \symb{Var}.\att{id} : \symb{Type}.\att{name} \} ; \\
& \symb{Decl}.\att{attrOut.isUsed[\symb{Var}]} \uparrow ~:=~ \text{false} ; \\
& \symb{Decl}.\att{attrOut.isInitialized[\symb{Var}]} \uparrow ~:=~ \text{false} ; \\ 
& \symb{Decl}.\att{valid} \uparrow ~:=~ \text{true} \big] \\
\end{array}$
\par\noindent

$\begin{array}{ll}
\textbf{b2.} & \symb{ObjInit}  :  ~\symb{Type}\$0~\symb{Var}~\textbf{= new}~\symb{Type}\$1~\symb{ArgList} \\
& \big[~\symb{ArgList}.\att{symTab} \downarrow ~:=~ \symb{ObjInit}.\att{symTab} \downarrow ;\\
& \symb{ObjInit}.\att{symtabOut} \uparrow ~:=~ \{ \symb{Var}.\att{id} : \symb{Type}.\att{name} \} ; \\
& \symb{ArgList}.\att{typeList} \downarrow ~:=~ \symb{Type}.\att{params} \uparrow;  \\
& \symb{ObjInit}.\att{attrOut.isInitialized[\symb{Var}]} \uparrow ~:=~ \text{true}; \\
& \symb{ObjInit}.\att{attrOut.isUsed[\symb{Var}]} \uparrow ~:=~ \text{false}; \\
& \symb{ObjInit}.\att{valid} \uparrow ~:=~ \symb{ArgList}.\att{valid} \uparrow; \\
& \ \wedge~ \symb{Type}\$0.\att{name} \uparrow ~:=~ \symb{Type}\$1.\att{name} \uparrow; \big] \\
\end{array}$

\par\noindent
$\begin{array}{ll}
\textbf{b3.} & \symb{Invoke}  :  \symb{Var}\$0~=~\symb{Var}\$1~\symb{Call}~\symb{InvokeMore} \\
& \big[~\symb{InvokeMore}.\att{symTab} \downarrow ~:=~ \symb{Invoke}.\att{symTab} \downarrow ;\\
& \symb{InvokeMore}.\att{exprType} \downarrow ~:=~ \symb{Call}.\att{retType} \uparrow; \\
& \symb{Call}.\att{attrIn} \downarrow ~:=~ \symb{Invoke}\$0.\att{attrIn} \downarrow; \\
& \symb{InvokeMore}.\att{attrIn} \downarrow~:=~\symb{Call}.\att{attrOut} \uparrow; \\
& \symb{Invoke}.\att{attrOut.isUsed[\symb{Var}\$0]} \uparrow ~:=~ \text{true}; \\
& \symb{Invoke}.\att{attrOut.isUsed[\symb{Var}\$1]} \uparrow ~:=~ \text{true}; \\
& \symb{Invoke}.\att{attrOut}\uparrow~:=~\symb{InvokeMore}.\att{attrOut} \uparrow; \\
& \symb{Invoke}.\att{valid} \uparrow ~:=~  \symb{InvokeMore}.\att{valid} \uparrow \\
& \  \wedge~ (\symb{InvokeMore}.\att{retType} \uparrow == \\
& \quad \quad \symb{Invoke}.\att{symTab} \downarrow [\symb{Var}\$0.\att{id} \uparrow]) \\
& \wedge~ \symb{Call}.\att{exprType} \uparrow ~==~ \\
&  \quad \quad \symb{Invoke}.\att{symTab} \downarrow[\symb{Var}\$1.\att{id} \uparrow]; \big] \\ 

\end{array}$

\par\noindent

$\begin{array}{ll}
 \textbf{b4a.} &  \symb{InvokeMore}\$0  :  ~\symb{Call}~\symb{InvokeMore}\$1 \\
& \big[~\symb{InvokeMore}\$1.\att{symTab} \downarrow ~:=~ \symb{InvokeMore}\$0.\att{symTab} \downarrow ;\\
& \symb{InvokeMore}\$1.\att{exprType} \downarrow ~:=~ \symb{Call}.\att{returnType} \uparrow; \\
& \symb{Call}.\att{symTab} \downarrow ~:=~ \symb{InvokeMore}\$0.\att{symTab} \downarrow ;\\
& \symb{InvokeMore}\$0.\att{retType} \uparrow ~:=~ \symb{InvokeMore}\$1.\att{retType} \uparrow; \\
& \symb{Call}.\att{attrIn} \downarrow ~:=~ \symb{InvokeMore}\$0.\att{attrIn} \downarrow; \\
& \symb{InvokeMore}\$1.\att{attrIn} \downarrow~:=~\symb{Call}.\att{attrOut} \uparrow; \\
& \symb{InvokeMoreOut}\$0.\att{attrIn} \uparrow~:=~\symb{InvokeMoreOut}\$1.\att{attrIn} \uparrow; \\
& \symb{InvokeMore}\$0.\att{valid} \uparrow ~:=~ \symb{Call}.\att{valid} \uparrow  \\
& \quad \wedge~ \symb{InvokeMore}\$1.\att{valid} \uparrow; \\
& \quad \wedge~ \symb{Call}.\att{exprType} \uparrow ~:=~ \symb{InvokeMore}\$1.\att{exprType} \downarrow; \big] \\
\end{array}$


$\begin{array}{ll}
\textbf{b4b.} & \symb{InvokeMore}  :  ~\epsilon \\
& \big[~\symb{InvokeMore}.\att{retType} \uparrow ~:=~ \symb{InvokeMore}.\att{exprType} \downarrow; \\
& \symb{InvokeMore}.\att{attrIn.itrVec} \uparrow = \text{(false, false)}; \\
& \symb{InvokeMore}.\att{valid} \uparrow ~:=~ \text{true} ; \big] \\
\end{array}$
\par\noindent
$\begin{array}{ll}
\textbf{b5} & \symb{Call}  :  \symb{Api}~\symb{ArgList} \\
& \big[~\symb{ArgList}.\att{symTab} \downarrow ~:=~ \symb{Call}.\att{symTab} \downarrow ;\\
& \symb{ArgList}.\att{typeList} \downarrow ~:=~ \symb{Api}.\att{params} \uparrow; \\
& \symb{Call}.\att{retType} \uparrow ~:=~ \symb{Api}.\att{retType} \uparrow] \\
& \symb{Api}.\att{attrIn} \downarrow ~:=~ \symb{Call}.\att{attrIn} \downarrow; \\
& \symb{Call}.\att{attrOut} \uparrow ~:=~ \symb{Api}.\att{attrOut} \uparrow; \\
& \symb{Call}.\att{exprType} \uparrow ~:=~ \symb{Api}.\att{exprType} \uparrow ); \\
& \symb{Call}.\att{valid} \uparrow ~:=~ \symb{ArgList}.\att{valid} \uparrow \big] \\
\end{array}$
\par\noindent

$\begin{array}{ll}
\textbf{b6a.} & \symb{ArgList}\$0 : ~\symb{Var}~\symb{ArgList}\$1~ \\
& \big[~\symb{ArgList}\$1.\att{symTab} \downarrow ~:=~ \symb{ArgList}\$0.\att{symTab} \downarrow ;\\
& \symb{ArgList}\$1.\att{typeList} \downarrow ~:=~ \symb{ArgList}\$0.\att{typeList}[1:] \downarrow; \\
& \symb{ArgList}\$1.\att{attrOut.isUsed[\symb{Var}]} \uparrow ~:=~ \text{true}; \\
& \symb{ArgList}\$0.\att{valid} \uparrow ~:=~ \symb{ArgList}\$1.\att{valid} \uparrow \\
& \ \ \ \wedge~ (\symb{ArgList}\$0.\att{symTab} \downarrow[\symb{Var}.\att{id} \uparrow]  \\
& \ \ \ \ \ \ \ == \symb{ArgList}\$0.\att{typeList}[0] \downarrow ); \big] \\
\end{array}$
\par\noindent
$\begin{array}{ll}
\textbf{b6b.} & \symb{ArgList}  : \epsilon \\
& \big[~\symb{ArgList}.\att{valid} \uparrow ~:=~ \symb{ArgList}.\att{typeList}.isEmpty() \uparrow; \big]\\
\end{array}$
\par\noindent
$\begin{array}{ll}
\textbf{b7.} & \symb{Return} : \textbf{return}~\symb{Var} \\
& \big[ ~\symb{Return}.\att{attrOut.retStmtGenerated} \uparrow ~::~\text{true} \\
& ~\symb{Return}.\att{valid} \uparrow ~:=~ ~\symb{Return}.\att{methodRetType} \downarrow == \\
& \quad \quad \symb{Return}.\att{symTab} \downarrow[\symb{Var}.\att{id} \uparrow] \big]; \\
\end{array}$

\par\noindent
\par\noindent

$\begin{array}{ll}
\textbf{c1.a.} & \symb{Stmt} : \symb{Branch} \\
& \big[~\symb{Branch}.\att{symTab} \downarrow ~:=~ \symb{Stmt}.\att{symTab} \downarrow ;\\
& \symb{Stmt}.\att{valid} \uparrow ~:=~ \symb{Branch}.\att{valid} \uparrow; \\
& \symb{Branch}.\att{attrIn} \downarrow ~:=~ \symb{Stmt}.\att{attrIn} \downarrow; \\
& \symb{Stmt}.\att{attrOut} \uparrow ~:=~ \symb{Branch}.\att{attrOut} \uparrow; \big] \\
\end{array}$
\par\noindent
$\begin{array}{ll}
\textbf{c1.b.} & \symb{Stmt} : \symb{Loop} \\
& \big[~\symb{Loop}.\att{symTab} \downarrow ~:=~ \symb{Stmt}.\att{symTab} \downarrow ;\\
& \symb{Stmt}.\att{valid} \uparrow ~:=~ \symb{Loop}.\att{valid} \uparrow; \\
& \symb{Loop}.\att{attrIn} \downarrow ~:=~ \symb{Stmt}.\att{attrIn} \downarrow; \\
& \symb{Stmt}.\att{attrOut} \uparrow ~:=~ \symb{Loop}.\att{attrOut} \uparrow; \big] \\
\end{array}$
\par\noindent
$\begin{array}{ll}
\textbf{c1.c.} & \symb{Stmt} :  \symb{Except}  \\
& \big[~\symb{Except}.\att{symTab} \downarrow ~:=~ \symb{Stmt}.\att{symTab} \downarrow ;\\
& \symb{Stmt}.\att{valid} \uparrow ~:=~ \symb{Except}.\att{valid} \uparrow; \\
& \symb{Except}.\att{attrIn} \downarrow ~:=~ \symb{Stmt}.\att{attrIn} \downarrow; \\
& \symb{Stmt}.\att{attrOut} \uparrow ~:=~ \symb{Except}.\att{attrOut} \uparrow; \big] \\
\end{array}$
\par\noindent
$\begin{array}{ll}
\textbf{c2.} & \symb{Branch} : \textbf{if}~\symb{Cond}~\textbf{then}~\symb{Stmt}\$1~\textbf{else}~\symb{Stmt}\$2 \\
& \big[~\symb{Cond}.\att{symTab} \downarrow ~:=~  \symb{Stmt}.\att{symTab} \downarrow ;\\
& \symb{Stmt}\$1.\att{symTab} \downarrow ~:=~  \symb{Cond}.\att{symtabOut} \uparrow ;\\
& \symb{Stmt}\$2.\att{symTab} \downarrow ~:=~ \symb{Cond}.\att{symtabOut} \uparrow ;\\
& \symb{Branch}.\att{valid} \uparrow ~:=~ \symb{Cond}.\att{valid} \uparrow \wedge~ \symb{Stmt}\$1.\att{valid} \uparrow  \\
& \quad \quad \quad \wedge~ \symb{Stmt}\$2.\att{valid} \uparrow;\\
& \symb{Cond}.\att{attrIn} \downarrow ~:=~ \symb{Branch}.\att{attrIn} \downarrow; \\
& \symb{Stmt}\$1.\att{attrIn} \downarrow ~:=~ \symb{Cond}.\att{attrIn} \downarrow; \\
& \symb{Stmt}\$2.\att{attrIn} \downarrow ~:=~ \symb{Cond}.\att{attrIn} \downarrow; \\
& \symb{Branch}.\att{attrOut} \uparrow ~:=~ \symb{Branch}\$1.\att{attrIn} \downarrow; \big] \\
\end{array}$
\par\noindent
$\begin{array}{ll}
\textbf{c3.} & \symb{Loop} : \textbf{while}~\symb{Cond}~\textbf{then}~\symb{Stmt} \\
& \big[~\symb{Cond}.\att{symTab} \downarrow ~:=~  \symb{Stmt}.\att{symTab} \downarrow ;\\
& \symb{Stmt}.\att{symTab} \downarrow ~:=~  \symb{Cond}.\att{symTabOut} \uparrow ;\\
& \symb{Loop}.\att{valid} \uparrow ~:=~ \symb{Cond}.\att{valid} \uparrow\wedge~  \symb{Stmt}.\att{valid} \uparrow; \\
& \symb{Cond}.\att{attrIn} \downarrow ~:=~ \symb{Loop}.\att{attrIn} \downarrow ; \\
& \symb{Stmt}.\att{attrIn} \downarrow ~:=~ \symb{Cond}.\att{attrOut} \uparrow; \\
& \symb{Loop}.\att{attrOut} \uparrow ~:=~ \symb{Loop}.\att{attrIn} \downarrow; \big] \\
\end{array}$
\par\noindent
$\begin{array}{ll}
\textbf{c4.} & \symb{Except} : \textbf{try}~\symb{Stmt}~\symb{Catch} \\
& \big[~\symb{Stmt}.\att{symTab} \downarrow ~:=~ \symb{Except}.\att{symTab} \downarrow ;\\
& \symb{Catch}.\att{symTab} \downarrow ~:=~ \symb{Stmt}.\att{symTabOut} \uparrow ;\\
& \symb{Except}.\att{valid} \uparrow ~:=~ \symb{Stmt}.\att{valid} \uparrow \wedge~ \symb{Catch}.\att{valid} \uparrow; \\
& \symb{Stmt}.\att{attrIn} \downarrow ~:=~ \symb{Except}.\att{attrIn} \downarrow ;\\
& \symb{Catch}.\att{attrIn} \downarrow ~:=~ \symb{Stmt}.\att{attrIn} \downarrow ;\\
& \symb{Except}.\att{attrOut} \uparrow ~:=~ \symb{Except}.\att{attrIn} \downarrow; \big] \\
\end{array}$
\par\noindent
$\begin{array}{ll}
\textbf{c5a.} & \symb{Catch}\$0 : \textbf{catch}(\symb{Type})~\symb{Stmt};~\symb{Catch}\$1~\\
& \big[~\symb{Catch}\$1.\att{symTab} \downarrow ~:=~ \symb{Catch}\$0.\att{symTab} \downarrow ;\\
& \symb{Stmt}.\att{symTab} \downarrow ~:=~ \symb{Catch}\$0.\att{symTab} \downarrow ;\\
& \symb{Stmt}.\att{attrIn} \downarrow ~:=~ \symb{Catch}\$0.\att{attrIn} \downarrow ;\\
& \symb{Catch}\$1.\att{attrIn} \downarrow ~:=~ \symb{Stmt}.\att{attrIn} \downarrow ;\\
& \symb{Catch}\$0.\att{attrOut} \uparrow ~:=~ \symb{Catch}\$1.\att{attrOut} \uparrow; \\
& ~\symb{Catch}\$0.\att{valid} \uparrow ~:=~ \symb{Stmt}.\att{valid} \uparrow \wedge \symb{Catch}\$1.\att{valid} \uparrow ; \big]\\
\end{array}$ \\

\par\noindent
$\begin{array}{ll}
\textbf{c5b.} & \symb{Catch} : ~\epsilon \\
& \big[~\symb{Catch}.\att{valid} \uparrow ~:=~ \text{true};  \symb{Catch}.\att{attrOut} \uparrow ~:=~ \phi \big]\\
\end{array}$ \\
\par\noindent

$\begin{array}{ll}
\textbf{c6.} & \symb{Cond} : \symb{Call} \\
& \big[~\symb{Call}.\att{symTab} \downarrow ~:=~ \symb{Cond}.\att{symTab} \downarrow ;\\
& \symb{Cond}.\att{valid} \uparrow ~:=~ \symb{Call}.\att{valid} \uparrow; \\
& \symb{Call}.\att{attrIn} \downarrow ~:=~ \symb{Cond}.\att{attrIn} \downarrow; \\
& \symb{Cond}.\att{attrOut} \uparrow ~:=~ \symb{Call}.\att{attrOut} \uparrow; \big] \\
\end{array}$
\endgroup 

$\begin{array}{ll}
\textbf{d1.} & \symb{Api}  :  \textbf{JAVA\_API\_CALL} \\
&\big[~ \symb{Api}.\att{name} \uparrow ~:=~ NAME;\\
& \symb{Api}.\att{params} \uparrow~:=~FORMAL\_PARAM\_LIST; \\
& \symb{Api}.\att{exprType} \uparrow~:=~TYPE; \\
& \symb{Api}.\att{retType}~\uparrow ~:=~ RET\_TYPE; \\
& \textnormal{if}(\symb{Api}.\att{name} == \textnormal{"hasNext"}) \\
& \quad \symb{Api}.\att{attrOut.itrVec} \uparrow ~:=~ (\text{true}, \text{false}); \\
& \textnormal{else if}(\symb{Api}.\att{name} == \textnormal{"next"}) \\
& \quad \symb{Api}.\att{attrOut.itrVec}[1] \uparrow ~:=~ \text{true}; \big] \\
\end{array}$
\par\noindent

$\begin{array}{ll}
\textbf{d2.} & \symb{Api}  :  \textbf{INTERNAL\_METHOD\_CALL} \\
&\big[~ \symb{Api}.\att{name} \uparrow ~:=~ NAME;\\
& \symb{Api}.\att{params} \uparrow~:=~FORMAL\_PARAM\_LIST; \\
& \symb{Api}.\att{exprType} \uparrow~:=~\epsilon; \\
& \symb{Api}.\att{retType}~\uparrow ~:=~ RET\_TYPE; \\
& \symb{Api}.\att{attrOut.itrVec} \uparrow ~:=~ \symb{Api}.\att{attrIn.itrVec} \downarrow;\big] \\

\end{array}$
\par\noindent

$\begin{array}{ll}
\textbf{d3.} & \symb{Type}  :  \textbf{JAVA\_TYPE} \\
& \big[~\symb{Type}.\att{name} \uparrow ~:=~ NAME \\
& \symb{Type}.\att{params} \uparrow ~:=~ FORMAL\_PARAM\_LIST; \big] \\
\end{array}$
\par\noindent
$\begin{array}{ll}
\textbf{d4.} & \symb{Var} : \textbf{VAR\_ID} \\ & \big[~\symb{Var}.\att{id} \uparrow ~:=~ ID\_NUMBER \big] \\
\end{array}$

\par\noindent
\par\noindent


\end{document}